\documentclass[conference]{IEEEtran}
\IEEEoverridecommandlockouts
\usepackage{cite}
\usepackage{amsmath,amssymb,amsfonts}
\usepackage{algorithmic}
\usepackage{graphicx}
\usepackage{textcomp}
\usepackage{xcolor}
\usepackage{hyperref}
\usepackage{array}
\usepackage{stfloats}
\usepackage{url}
\usepackage{verbatim}
\usepackage{xstring}
\usepackage{subfigure}
\usepackage{epsfig} 
\usepackage{times} 
\usepackage{mathtools}
\usepackage{multirow}
\usepackage{rotating}
\usepackage{mwe}
\usepackage{float}
\usepackage{mathtools}
\usepackage{blkarray, bigstrut} %
\usepackage{bm}

\makeatletter
\def\ps@IEEEtitlepagestyle{%
  \def\@oddfoot{\mycopyrightnotice}%
  \def\@oddhead{\hbox{}\@IEEEheaderstyle\leftmark\hfil\thepage}\relax
  \def\@evenhead{\@IEEEheaderstyle\thepage\hfil\leftmark\hbox{}}\relax
  \def\@evenfoot{}%
}
\def\mycopyrightnotice{%
  \begin{minipage}{\textwidth}
  \centering \scriptsize
  Copyright~\copyright~2024 IEEE. Personal use of this material is permitted. Permission from IEEE must be obtained for all other uses, in any current or future media, including\\reprinting/republishing this material for advertising or promotional purposes, creating new collective works, for resale or redistribution to servers or lists, or reuse of any copyrighted component of this work in other works by sending a request to pubs-permissions@ieee.org.
  \end{minipage}
}
\makeatother

\def\BibTeX{{\rm B\kern-.05em{\sc i\kern-.025em b}\kern-.08em
    T\kern-.1667em\lower.7ex\hbox{E}\kern-.125emX}}
\begin{document}

\title{Enhancing State Estimator for Autonomous Racing : Leveraging Multi-modal System and Managing Computing Resources




\thanks{\textsuperscript{1} School of Electrical Engineering, Korea Advanced Institute of Science and Technologies (KAIST), Daejeon, Republic of Korea
        {\texttt{\{lee.dk, menu107, ryuchanhoe, sw.nah, seongwoo.moon, hcshim\}@kaist.ac.kr}}}
\thanks{\textsuperscript{*} Corresponding Author}
\thanks{This work was supported by the Technology Innovation Program (RS-2023-00256794, Development of drone-robot cooperative multimodal delivery technology for cargo with a maximum weight of 40kg in urban areas) funded By the Ministry of Trade, Industry \& Energy(MOTIE, Korea).}
}

\author{\IEEEauthorblockN{Daegyu Lee\textsuperscript{1}, Hyunwoo Nam\textsuperscript{1}, Chanhoe Ryu\textsuperscript{1}, Sungwon Nah\textsuperscript{1}, Seongwoo Moon\textsuperscript{1} and D. Hyunchul Shim\textsuperscript{1*}}
}

\maketitle

\begin{abstract}
This paper introduces an approach that enhances the state estimator for high-speed autonomous race cars, addressing challenges from unreliable measurements, localization failures, and computing resource management.
The proposed robust localization system utilizes a Bayesian-based probabilistic approach to evaluate multimodal measurements, ensuring the use of credible data for accurate and reliable localization, even in harsh racing conditions.
To tackle potential localization failures, we present a resilient navigation system which enables the race car to continue track-following by leveraging direct perception information in planning and execution, ensuring continuous performance despite localization disruptions.
In addition, efficient computing is critical to avoid overload and system failure. 
Hence, we optimize computing resources using an efficient LiDAR-based state estimation method.
Leveraging CUDA programming and GPU acceleration, we perform nearest points search and covariance computation efficiently, overcoming CPU bottlenecks.
\par
Simulation and real-world tests validate the system's performance and resilience. The proposed approach successfully recovers from failures, effectively preventing accidents and ensuring safety of the car.
\end{abstract}


\section{Introduction}
High-performance robotics racing is emerging as a powerful platform for driving development and showcasing state-of-the-art capabilities in autonomous systems. 
The first Indy Autonomous Challenge (IAC) started in October 2021, utilizing the Dallara IL-15 IAC race car, equipped with six mono-cameras, three radars, three LiDARs, and two real-time kinematic (RTK) GPS (Global Positioning System). 
Since then, four races have taken place, including two at the Consumer Electronics Show (CES) in Las Vegas, one in Texas in November 2022, and the most recent one in Monza, Italy. 
These competitions emphasize the development of software capable of effectively handling edge cases in autonomous driving, even in scenarios involving speeds of up to 300 $km/h$.
\par

Parallel to these developments, a new autonomous racing competition, the Hyundai Autonomous Challenge (HAC), is emerging in South Korea, featuring Hyundai Motors' electric car, the IONIQ 5. 
While each team can customize the sensors for their vehicles, there is a common trend toward using LiDAR and GPS/inertial navigation system (INS) in the high-speed navigation algorithm. The primary goal of this competition, similar to the IAC, is to stimulate the advancement of autonomous driving technology.


\begin{figure}[!t]
    \centering
    \subfigure[]{
        \includegraphics[width=0.45\columnwidth]{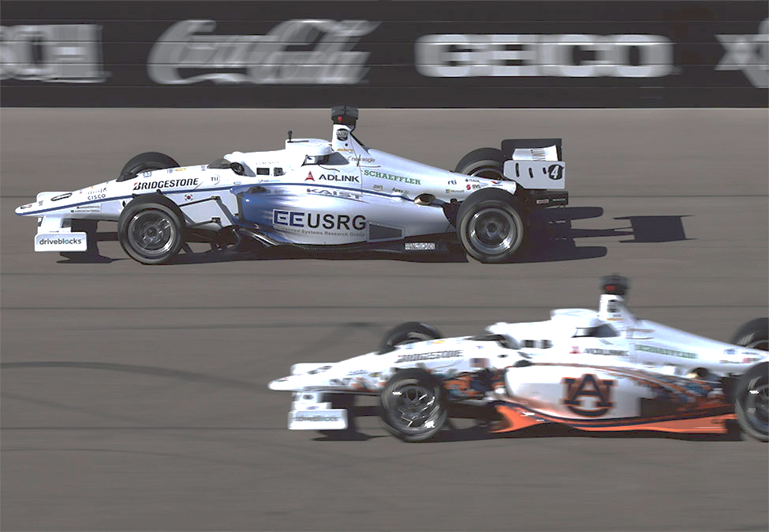}
        \label{fig:main_1}
    }
    \subfigure[]{
        \includegraphics[width=0.45\columnwidth]{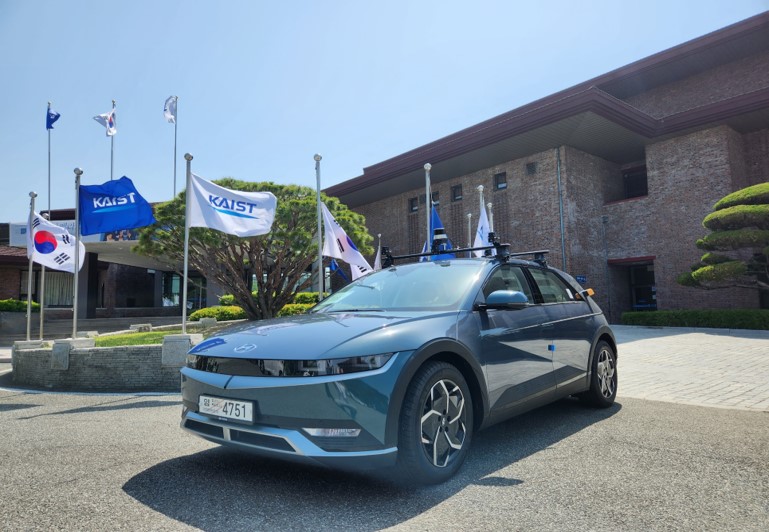}
        \label{fig:main_2}
    }
    \label{fig:main}
    \caption[Dallara AV-21 of Team KAIST at the Las Vegas Motor Speedway (LVMS).]{
    (a)Dallara AV-21 of Team KAIST at the Las Vegas Motor Speedway (LVMS)
    (b)IONIQ 5 of Team KAIST for HAC 2023.}
\end{figure}

However, the deployment of high-speed race cars revealed a significant correlation between accidents and issues with navigation systems. 
The majority of participating teams faced challenges arising from unreliable sensor measurements, particularly impacting localization performance. 
The GPS, a critical sensor for localization, posed significant challenges during high-speed driving, as vibrations often led to drifts in position solutions, resulting in serious accidents.
\par
In addition to these challenges, common autonomous vehicles encounter constraints in installing additional computing devices, necessitating efficient management of computing resources. 
To address these critical issues and enhance system robustness, a comprehensive approach to localization becomes imperative. 
Multimodal sensing, involving the integration of diverse sensors, offers a viable solution. In this study, GPS/INS and LiDARs serve as our primary sensors. 
Leveraging this multimodal sensing system, we present a resilient and efficient localization framework, encompassing the following key components:
\begin{itemize}
\item Real-time identification of measurement reliability is achieved through a probabilistic Bayesian approach. This facilitates selective updates by filtering out unreliable measurements.
\item An efficient LiDAR-based state estimation method, incorporating two core concepts: computing efficient scan matching for correction and LiDAR odometry for prediction, both accelerated by CUDA.
\item A resilient navigation system that utilizes direct LiDAR measurements to maintain safe operation of the system, even in the event of localization failures.
\end{itemize}
\par
Throughout a series of computer simulations and experiments on real racetracks, it is validated that the proposed robust localization algorithm and the efficient navigation module are capable of maintaining high-speed racing even when all GPS sensors are temporarily deteriorated.
Moreover, this research not only addresses the challenges observed in high-speed race car deployments but also contributes valuable insights to the broader domain of autonomous vehicle navigation under challenging conditions, emphasizing the importance of efficient computing resource management.
\par
The remainder of the paper is organized as follows. Section \ref{sec:related} introduces related studies. 
Section \ref{sec:methods} describes our high-speed, resilient state estimator leveraging GPS/INS and LiDAR.
The experimental results, including simulated and real-world tests, are discussed in Section \ref{sec:results}. 
Finally, Section \ref{sec:conclusion} concludes this paper, and potential improvements and future works are explored in Section \ref{sec:discussion}.

\section{Related works}
\label{sec:related}
\subsection{Localization based on multi-sensor fusion}
The key challenge of navigation is adapting, learning, and recovering from failures \cite{yang2018grand}.
To address this issue, multi-sensor fusion has been studied.
A challenge is to integrate multiple sensors and distinguish between accurate and unreliable ones.
Tightly coupled navigation with global navigation satellite system (GNSS), LiDAR, inertial measurement unit (IMU) is proposed in \cite{gao2015ins}. 
In addition, Soloviev \textit{et al}. \cite{soloviev2007tight} presented tight coupling of LiDAR and IMU without GPS. 
However, these applications are deployed on the limited scenes due to 2D laser scanner.
Recently, robust and precise localization system is proposed \cite{wan2018robust}. 
In the study, they evaluated the localization system in various city scenes by integrating 3-D LiDAR, GPS, and IMU. However, only one sensor of each type was used. 
\par
In most open sky environments, GPS plays an important role in localization by providing accurate global position, but their performance might be impacted in environments with occlusions, multi-path or radio interferences. 
Therefore, significant studies on LiDAR-based multi-sensor fusion for GPS-denied environments have been conducted \cite{xu2022review}, deploying robots in campus-scale environments \cite{li2020multi} or expanding to city-scale urban areas \cite{lee2022design}.
In this study, we utilize GPS/INS and LiDAR for high-speed driving systems, and present research on multimodal measurement utilizing multiple GPS and LiDAR.

\subsection{Perception based navigation system}
Perception based navigation systems offer high accuracy and flexibility in dynamic environments, particularly by enabling continued navigation even in the absence of GPS signals. 
Wall-following navigation based on LiDAR sensors has demonstrated promising results, as observed in tunnel scenarios \cite{che2022wall}, narrow small-scaled environments \cite{csaba2018mobil}, and in scenarios involving columns or dynamic obstacles \cite{joo2022wall}.
In these studies, a wall following algorithm was introduced for mobile platforms, primarily using 2D LiDAR sensors. 
However, our system employs 3D LiDAR, requiring ground filtering to distinguish between walls and the ground. 
Additionally, when applied to high-speed autonomous vehicles on a race track, wall following approach presents unique challenges.
\par
RGB images can also be utilized for perception-based navigation system \cite{wang2019improved}.
To improve image-based approach, hybrid methods that combine point-cloud and images have been proposed in recent studies \cite{zeilinger2017design,  kabzan2020amz}. 
Recently, deep perception-based approaches for autonomous race cars have been studied, utilizing end-to-end systems to learn control policies from on-board observations \cite{pan2020imitation}. 
Collaborative efforts with sim-to-real approaches have also been explored \cite{balaji2019deepracer}, along with studies on handling cross-domain model transfer \cite{wurman2022challenges}. 
However, there is an ongoing need to ensure that the output is reliable and free from fatal errors in real-world scenarios.

\subsection{Autonomous racing}
In the past few years, scale RC cars have been used to 
investigate algorithms for autonomous racing experimentally \cite{liniger2015optimization, babu2020f1tenth}.
A comprehensive overview of the current autonomous racing platforms, emphasizing the software-hardware co-evolution to the current stage, is presented in \cite{betz2022autonomous}.
After a series of IAC races, each team presented its full stack of autonomous systems and approaches in \cite{betz2022tum, spisak2022robust, hartmann2022competitive}.
Furthermore, the TUM Autonomous Motorsport team has introduced planning and control modules, employing strategies such as minimizing curvature trajectory \cite{heilmeier2019minimum} and optimizing time-optimal trajectories \cite{christ2021time}. 
They have also explored model-based approaches \cite{wischnewski2022tube, wischnewski2022indy}, incorporating vehicle-model estimation \cite{hermansdorfer2019concept} and minimizing lap times considering road conditions \cite{herrmann2020minimum}.
In addition, when autonomous racing becomes analogous to human-driven racing competition, 
behavior planning is crucial for performing overtaking maneuvers or decision-making, and a reinforcement learning-based approach has been recognized as significant in this context \cite{niu2020two, song2021autonomous}.
Hence, we studied the game-based predictor to predict future trajectories of racing competitors for overtaking or decision-making in autonomous racing \cite{jung2021game}.

\section{Methods}
\label{sec:methods}
\begin{figure}[t]
    \centering
    \includegraphics[width=1.0\columnwidth]{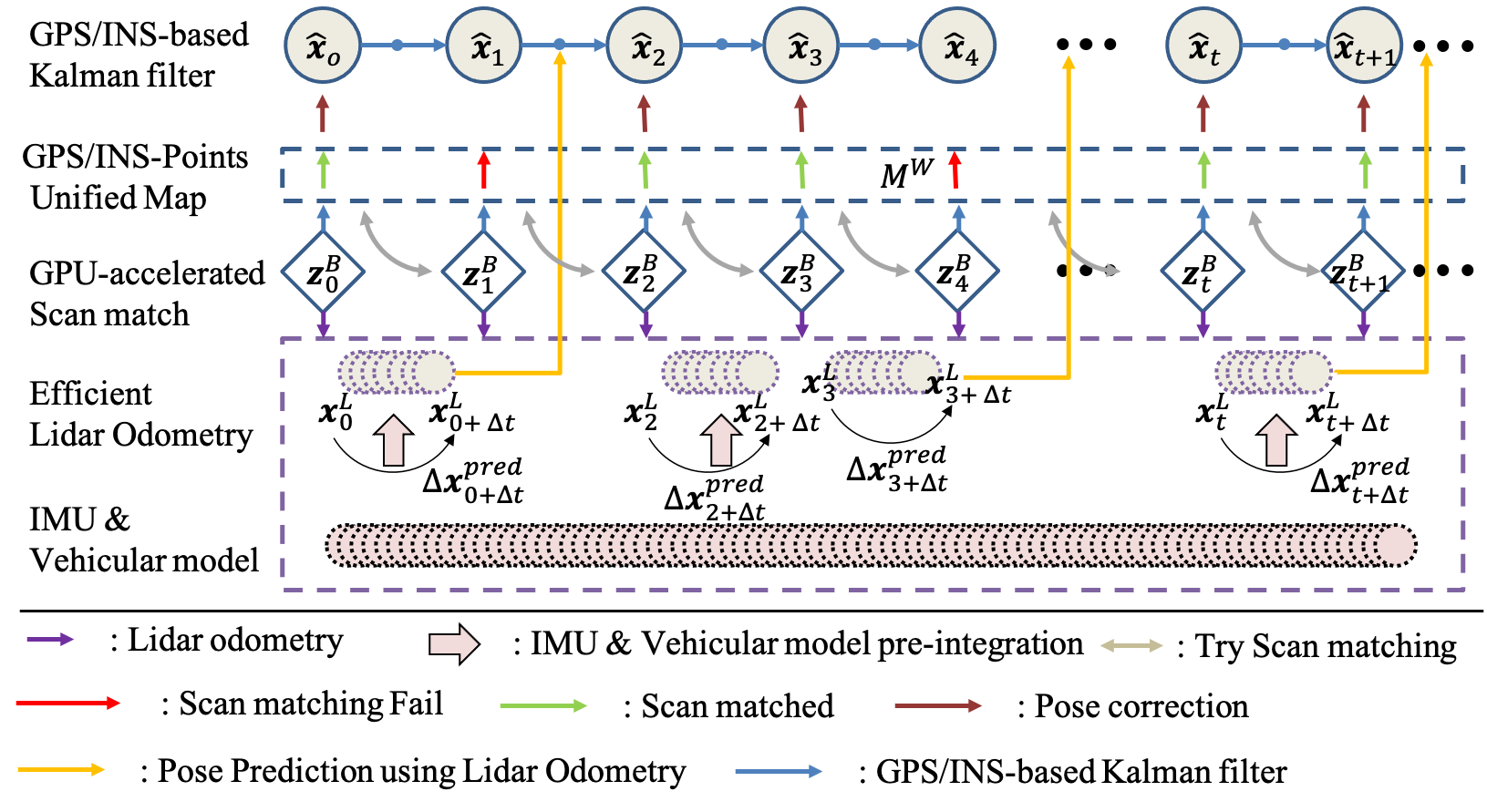}
    \caption[Proposed state estimator for high-speed race car]{Proposed state estimator for high-speed race car}
    \label{fig:state_estimator}
\end{figure}

To integrate multimodal measurement system, our full state estimatior is designed as shown in Fig. \ref{fig:state_estimator}.
We integrate a Kalman filter-based GPS/INS method with a LiDAR-aided state estimation algorithm. 

\subsection{Multimodal measurement fusion Kalman filter}
For multimodal sensor fusion, we use an extended Kalman filter algorithm. 
Although Kalman filter enables us to combine information from multiple GPS units and LiDAR sensors, its performance deteriorates rapidly when unreliable measurements are fed to the filter. 
In order to prevent such situations, we identify measurement reliability in real-time through a probabilistic Bayesian approach, and selectively update the multimodal measurements.
\par
We propose our multimodal measurement fusion Kalman filter algorithm to deal with GPS measurement noise due to strong vibrations. 
To obtain global positioning using LiDAR, we create a pre-built map of a race track and apply a scan-matching algorithm.

\subsubsection{Vehicle state estimation formulation}
The Kalman filtering process of time-varying stochastic control system with $k$ multiple measurements is given by
\begin{equation}
\begin{aligned}
    & \mathbf{x}_{t} = \mathbf{F}_{t-1}\mathbf{x}_{t-1} + \mathbf{B}_{t-1}\mathbf{u}_{t-1} + \mathbf{q}_{t}, \\
    & \mathbf{y}_{t}^{k} = \mathbf{H}_{t-1}^{k}\mathbf{x}_{t-1} + \mathbf{W}_{t}, 
\end{aligned}
\label{eq_kalman_basic}
\end{equation}
where $\mathbf{x}_{t} \in \mathbb{R}^n$ is the state, $\mathbf{y}_{t}^{k} \in \mathbb{R}^m$ is the $k$-th measurement, and $\mathbf{u}_{t} \in \mathbb{R}^l$ is the control input. $\mathbf{q}_{t} \sim \mathcal{N}(0, \mathbf{Q})$ and $\mathbf{W}_{t} \sim \mathcal{N}(0, \mathbf{R})$ are an additive white Gaussian noise (AWGN). 
Specifically, to achieve time-efficient state estimation with the available computation hardware, we used extended Kalman filter (EKF) with constant turn rate and velocity (CTRV) model \cite{tao_watanabe_yamada_takada_2021}.
Therefore, we denote the vehicle state as
\begin{equation}
\begin{aligned}
    & \mathbf{x}_{t} = [x_t, y_t, \theta_t, b_t]^T, \\
    & \mathbf{u}_{t} = [v_{x,t}, w_{z,t}]^T, \\
    & \mathbf{y}^k_{t} = [\mathcal{Z}_{x,t}^k, \mathcal{Z}_{y,t}^k, \mathcal{Z}_{\theta, t}^k]^T, 
\end{aligned}
\label{eq_state}
\end{equation}
where $\{x_t, y_t, \theta_t\} \in SE(2)$ are vehicle's coordinate, heading, and $b_t$ is yaw bias.
Control input $\{v_{x,t}, w_{z,t}\}$ are longitudinal velocity and yaw velocity, respectively, and they are obtained from the vehicle electronics and IMU.
We assume the evolution of each element in Eq. \ref{eq_state} is described by and vehicle's nonlinear kinematics model as 
\begin{equation}
\begin{aligned}
    x_{t} & = x_{t-1} + v_{x,t-1}  cos(\theta_{t-1} + b_{t-1})  dt,  \\
          & = x_{t-1} + \psi_{t-1}  dt\\
    y_{t} & = y_{t-1} + v_{x,t-1}  sin(\theta_{t-1} + b_{t-1})  dt,\\
          & = y_{t-1} + \phi_{t-1}  dt\\
    \theta_{t} & = \theta_{t-1} + w_{z,t-1}  dt, \\
    b_{t} & = b_{t-1}, v_{x,t} = v_{x,t-1}, w_{z,t} = w_{z,t-1}.
\end{aligned}
\label{eq_state_specific}
\end{equation}
To extend linear model with Eq. \ref{eq_kalman_basic}, we obtain the discrete-time evolution model with time-dependent linearized matrix $\mathbf{A}_t$ as 
\begin{equation}
\begin{aligned}
    & \mathbf{x}_{t} = \mathbf{A}_{t} \cdot \mathbf{x}_{t-1} + \mathbf{q}_{t},
\end{aligned}
\label{eq_linear_state}
\end{equation}
where $\mathbf{A}_t$ can be represented as
\begin{equation}
\begin{aligned}
    \mathbf{A}_{t} & = \\
    & \begin{bmatrix} 
    1 & 0 & \dot{\psi}_{t-1}\cdot dt & \dot{\psi}_{t-1}\cdot dt & {\psi}_{t-1} / v_{x,t-1} \cdot dt & 0\\
    0 & 1 & \dot{\phi}_{t-1}\cdot dt & \dot{\phi}_{t-1}\cdot dt & {\phi}_{t-1} / v_{x,t-1} \cdot dt & 0\\
    0 & 0 & 1 & 0 & 0 & dt \\
    0 & 0 & 0 & 1 & 0 & 0 \\
    0 & 0 & 0 & 0 & 1 & dt \\
    0 & 0 & 0 & 0 & 0 & 1 \\
\end{bmatrix}.
\end{aligned}
\label{eq_state_extend}
\end{equation}

\subsubsection{Problem definition}
Our goal is to estimate vehicle state with $k$ observation model to obtain the $k$ observations vector as 
\begin{equation}
\begin{aligned}
    & \mathbf{y}^k_{t} = \mathbf{h}^{k}(\mathbf{x}_t) + \mathbf{W}_t. 
\end{aligned}
\label{eq_state_problem_define}
\end{equation}
We can address our localization problem as the posterior probability in terms of the prior likelihood \cite{charles2017kalman}.
Our problem definition starts with optimizing the overall state of $\hat{\mathbf{x}}_t$ with multiple observations $\mathbf{y}_{t}^{k}$ : 
\begin{equation}
\begin{aligned}
    & \{\hat{\mathbf{x}}_t\}_{t \in [1,t]} = \arg\max[\prod_{i=0}^{t} p(\mathbf{x}_{i}|\mathbf{x}_{i-1})p(\mathbf{y}_{i}^{k}|\mathbf{x}_{i})p(\mathbf{y}_{0}^{k}|\mathbf{x}_{0})p(\mathbf{x}_{0})].
\end{aligned}
\label{eq_kalman_1}
\end{equation}
We define our localization problem as finding an optimal solution at the current time $t$; a marginalization can be represented as follows:
\begin{equation}
\begin{aligned}
    \hat{\mathbf{x}}_t = & \arg\max_{\mathbf{x}_{t}}[\int p(\mathbf{y}_{i}^{k}|\mathbf{x}_{t})p(\mathbf{x}_{t}|\mathbf{x}_{t-1}) \\ 
    & \prod_{i=0}^{t} p(\mathbf{x}_{i}|\mathbf{x}_{i-1})p(\mathbf{y}_{i}^{k}|\mathbf{x}_{i})p(\mathbf{y}_{0}^{k}|\mathbf{x}_{0})p(\mathbf{x}_{0})d\{\mathbf{x}_{l}\}_{l \in [1,t-1]}],
\end{aligned}
\label{eq_kalman_2}
\end{equation}
where we assume each variable as independent and identically distributed.
We further simplify Eq. \ref{eq_kalman_2} as
\begin{equation}
\begin{aligned}
    \hat{\mathbf{x}}_t \propto & \arg\max_{\mathbf{x}_{i}}[\int p(\mathbf{y}_{i}^{k}|\mathbf{x}_{i})p(\mathbf{x}_{t}|\mathbf{x}_{t-1})d\{\mathbf{x}_{l}\}_{l \in [1,t-1]}].
\end{aligned}
\label{eq_kalman_3}
\end{equation}
Using Bayes' theorem, we propose a hyper-parameter $\Theta$ for localization to conditionally update the measurement as follows: 
\begin{equation}
\begin{aligned}
    \hat{\mathbf{x}}_t \propto & \arg\max_{\Theta}[\int p(\mathbf{y}_{i}^{k}|\Theta^{k})p(\Theta^{k}|\mathbf{x}_{i})p(\mathbf{x}_{t}|\mathbf{x}_{t-1})d\{\mathbf{x}_{l}\}_{l \in [1,t-1]}].
\end{aligned}
\label{eq_kalman_4}
\end{equation}
Intuitively, it is the likelihood of the $k$-th measurement $\mathbf{y}_{t}^{k}$ over all possible parameter values, weighted by the prior $p(\Theta^{k}|\mathbf{x}_{i})$.
If all localization hyper-parameter $p(\Theta^{k})$ assign high probability to the $k$-th measurement data, then this is a reasonable criterion for multiple measurements.
We define our multimodal measurement fusion algorithm for high-speed localization as finding $p(\mathbf{y}_{i}^{k}|\Theta^{k})$ and $p(\Theta^{k}|\mathbf{x}_{i})$.
Therefore, we determine a poor signal quality computing hyper-parameter $\Theta$ to prevent using weak measurements to correct the Kalman filter. 
\begin{figure}[t]
    \centering
    \includegraphics[width=0.75\columnwidth]{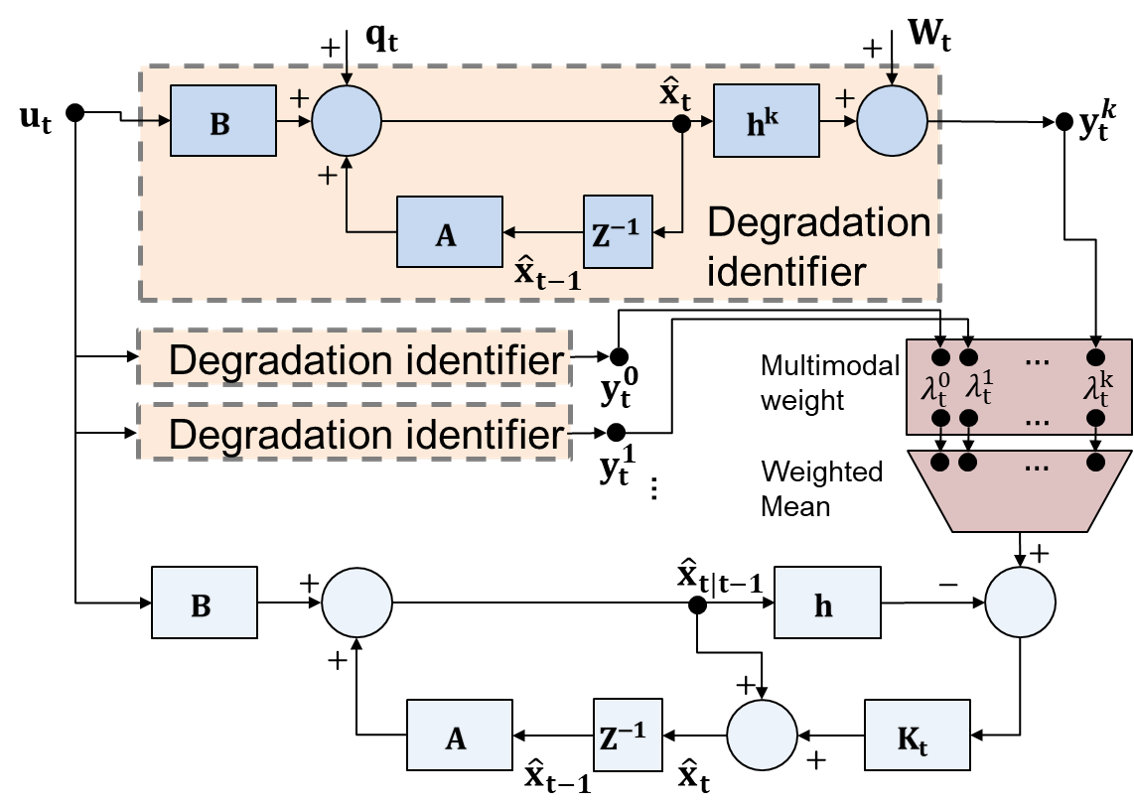}
    \caption[Diagram of Kalman filter for multimodal measurement and degradation identification]{Diagram of Kalman filter for multimodal measurement and degradation identification}
    \label{fig:diagram}
\end{figure}
Our proposed Kalman filter for multimodal measurement and degradation identification is described in Fig. \ref{fig:diagram}.

\subsubsection{Bayesian-based multimodal measurement update}
At a correction step of the Kalman filter, we propose a degradation identification method with a novel hyper-parameter $p(\Theta^{k})$ derived from Bayesian decision theory. 
We assume $p(\Theta^{k}|\mathbf{x}_{t})$ using the Mahalanobis distance $\Delta_{k}$ \cite{de2000mahalanobis} as follows:
\begin{equation}
\begin{aligned}  
    & p(\Theta^{k}|\mathbf{x}_{t}) 
    \triangleq 1 - 1 / (1 + exp(-\Delta_{k})),
\end{aligned}
\label{eq_sigmoid_mahala}
\end{equation}
where $\Delta_{k}$ pass through the sigmoid function.
Specifically, $\Delta_{k}$ can be obtained as 
\begin{equation}
\begin{aligned}
    & \Delta_{k}
    = (\mathbf{x}_{t-1} - \mathbf{y}^{k})^T\Sigma^{-1}_{k}(\mathbf{x}_{t-1} - \mathbf{y}^{k}),
\end{aligned}
\label{eq_mahala}
\end{equation}
where $\mathbf{y}^{k}$ is the $k$-th measurement and $\Sigma_{k}^{-1} = \Lambda_k$ is the $k$-th measurement precision matrix.
$\Sigma_{k}$ is updated with AWGN $\mathbf{n}_t$ and covariance value from measurement.
More precisely, 
We interpret $\Delta_{k}$ as a Euclidean distance in an orthogonally transformed new coordinate frame to identify whether $\mathbf{y}^{k}$ is a reasonable input or not. 

We propose a novel method for updating measurements, which is based on Bayesian decision theory.
Let $p(\mathbf{y}_{i}^{k}|\Theta^{k}, \mathbf{x}_{i})$ defined conditional to $\Delta_{k}$ as follows:
\begin{equation}
    \begin{aligned}
        p(\mathbf{y}_{i}^{k}|\Theta^{k}, \mathbf{x}_{i})  
        = 
        \begin{cases}
        &  \mathbf{y}^0 \hfill \text{ if } \forall\Delta_{k} \leq \epsilon \\
        &  \lambda_{i} \cdot \mathbf{y}^i \hfill \text{ if } \forall\Delta_{k} > \epsilon \text{ and } \forall\Delta_{k} \leq \delta \\
        &  \mathbf{y}^{k} \hfill \text{ if } \Delta_{k} \leq \delta \text{ and } \forall\Delta_{ \sim k} > \delta \\
        & \lambda_{reject} \hfill \text{ if } \forall\Delta_{k} > \delta, \\    
        \end{cases}
    \end{aligned}
    \label{eq_conditional_measure}
\end{equation}
where $\lambda_{k} = 1 - \Delta_k / \sum(\Delta_i)$, $\delta$, and $\epsilon$ are the measurement update weight, hyper-parameter to determine qualified measurement, and error($\epsilon << \delta$), respectively.
If all  $\Delta_{k}$ is less than $\epsilon$, we use one of the measurements because it is already qualified.
If all $\Delta_{k}$ is less than $\delta$, we implement the weighted sum of measurement, assuming that all the measurement is in reasonable condition.
In contrast, if $\Delta_{k}$ is less than $\delta$, but $\Delta_{\sim k}$ is greater than $\delta$, we use the only one feasible measurement, $\mathbf{y}^k$.
Here, $\sim k$ in $\Delta_{\sim k}$ denotes complement set of $\Delta_{k}$.  
Lastly, if all the distance is unsuitable for measurement updates, a warning alarm is raised to be ready for the resilient localization system to be discussed in the next section \ref{sec:sub_wall_detection}.
If the new measurement status is rejected, we update our state only depending on control input $\mathbf{u}_t$ without utilizing $\mathbf{y}^k$ at all.

\subsection{Unified frame map generation}
\begin{figure}[!t]
    \centering
    \subfigure[]{
        \includegraphics[width=1.0\columnwidth]{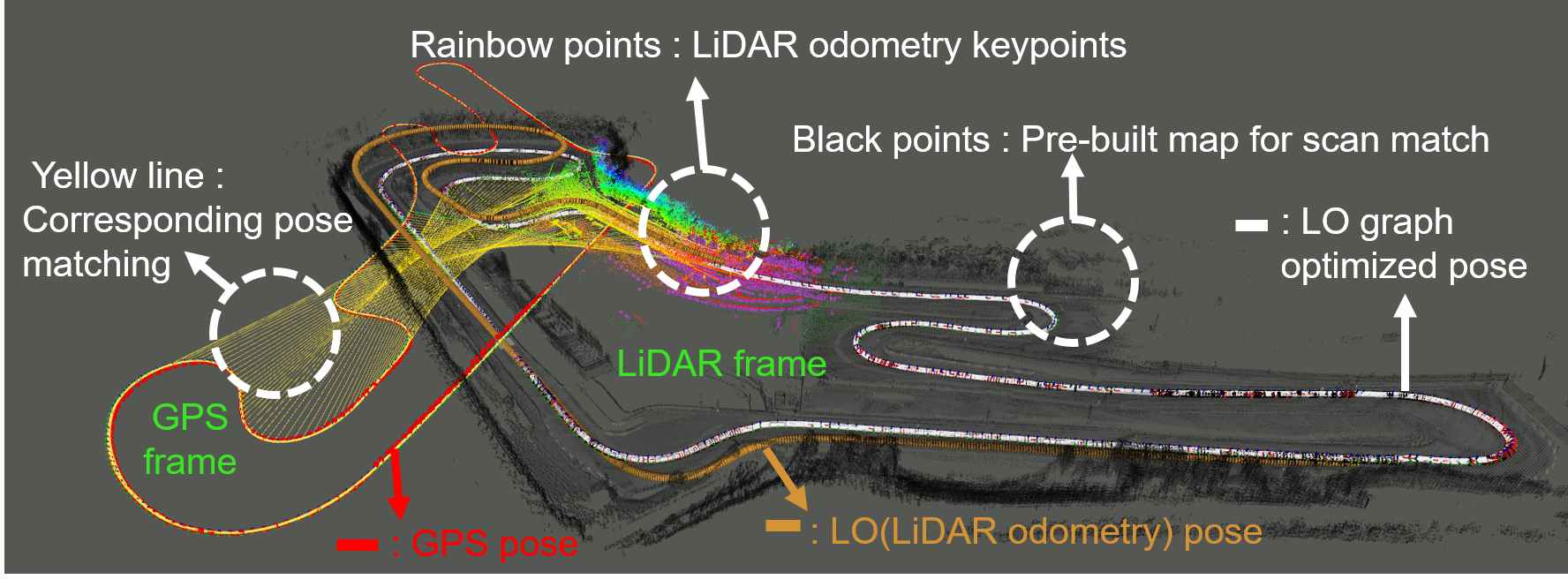}
        \label{fig:multi_frames}
    }
    \subfigure[]{
    \includegraphics[width=1.0\columnwidth]{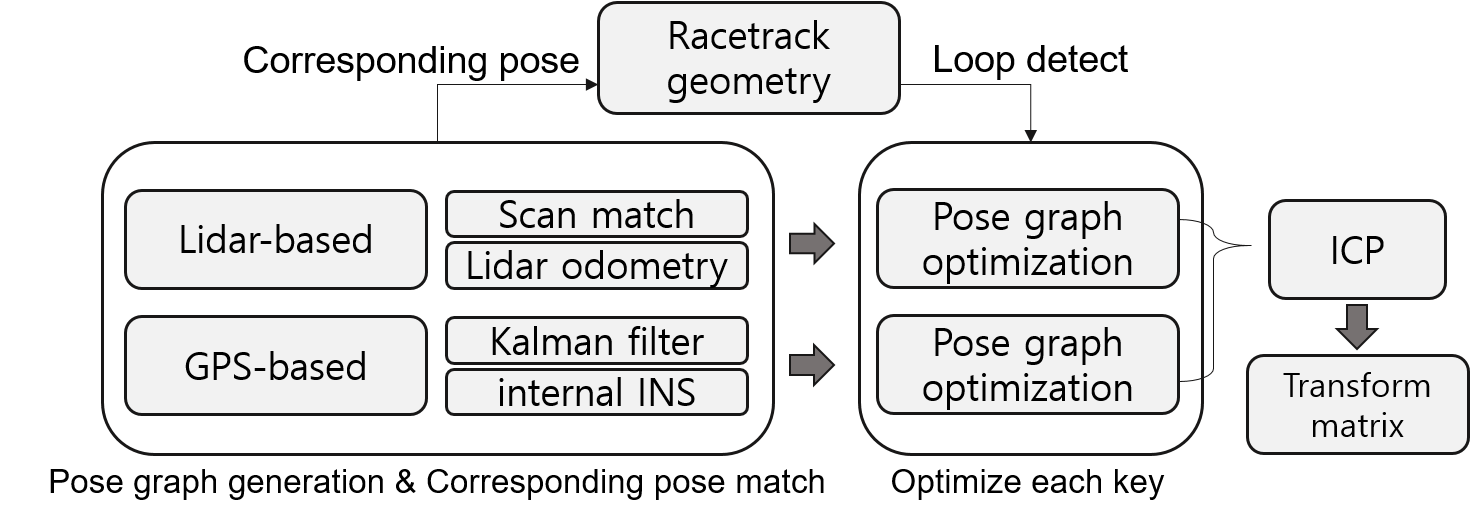}
    \label{fig:pose_graph_opt}
    } 
    \caption[This should be changed to clear plot.]{
    The corresponding pose graph is visualized, showcasing the relationship between the LiDAR point cloud and GPS/INS frames of National Automobile Racetrack of Monza, Italy.
    }
\end{figure}

In a multimodal system, establishing accurate transformations between multiple global frames is crucial, as depicted in fig. \ref{fig:multi_frames}.
Georeferencing plays a key role in this process \cite{hackeloeer2014georeferencing}.
Seamlessly to utilize a LiDAR scan-matched pose $\mathbf{y}^k_{scan} \in \mathbf{y}^k$ as a multimodal measurement, generating a unified map based on GPS frames is more intuitive, given that $\mathbf{y}^k_{scan}$ can be directly integrated.
\par
To facilitate this process, we employ pose-graph optimization, crafted to construct a unified 3-D map denoted as $\mathbf{M}^W$. 
This is accomplished by minimizing the spatial separation between corresponding poses within the map. 
Notably, this section introduces a unique pose-graph optimization-based frame unification method tailored specifically for the racetrack environment. 
The outlined approach is depicted in fig. \ref{fig:pose_graph_opt}.
\par

Using GTSAM framework \cite{dellaert2012factor}, we update factor graphs using the GPS/INS key pose $\mathbf{x}^G_i \in \mathbf{x}_i$ and LiDAR key pose $\mathbf{x}^L_i \in \mathbf{x}_i$.
Therefore, the factor graphs can be represented as 
\begin{equation}
\begin{aligned}
    & f(\mathbf{x}^G_0, \mathbf{x}^G_1, ..., \mathbf{x}^G_{n-1}) = \prod f_i(\mathcal{X}_i^G), \\ 
    & f(\mathbf{x}^L_0, \mathbf{x}^L_1, ..., \mathbf{x}^L_{n-1}) = \prod f_i(\mathcal{X}_i^L), 
\end{aligned}
\label{eq_factor_grpah}
\end{equation}
where $f_i$ indicates the connectivity of a factor graph's each factor. 
\par
To update the GPS/INS-based pose graph $f_i(\mathcal{X}_i^G)$ and LiDAR-based pose graph $f_i(\mathcal{X}_i^L)$, we process each frame individually. The update equations for GPS/INS-based odometry measurements are:
\begin{equation}
\begin{alignedat}{2}
& f_0(\mathcal{X}_0^G) = f_0(\mathbf{x}_1^G) , \quad && \text{for} \ i = 0 , \\
& f_i(\mathcal{X}_i^G) = f_i(\mathbf{x}_i^G, \mathbf{x}_{i+1}^G; \mathbf{o}_i^G) , \quad && \text{for} \ i > 0 .
\end{alignedat}
\label{eq_gps_factor}
\end{equation}
For LiDAR odometry measurements:
\begin{equation}
\begin{alignedat}{2}
& f_0(\mathcal{X}_0^L) = f_0(\mathbf{x}_1^L) , \quad && \text{for} \ i = 0 , \\
& f_i(\mathcal{X}_i^L) = f_i(\mathbf{x}_i^L, \mathbf{x}_{i+1}^L; \mathbf{o}_i^L) , \quad && \text{for} \ i > 0 .
\end{alignedat}
\label{eq_lidar_factor}
\end{equation}
Here, $\mathbf{o}_i^G$ represents GPS/INS-based odometry measurements, and $\mathbf{o}_i^L$ represents LiDAR odometry measurements. 
For $i = 0$, the initial pose is updated using unary factors $f_0(\mathbf{x}_1^G)$ and $f_0(\mathbf{x}_1^L)$ for GPS/INS and LiDAR respectively. 
For subsequent poses $(i > 0)$, the updates depend on factors $f_i(\mathbf{x}i^G, \mathbf{x}{i+1}^G; \mathbf{o}_i^G)$ and $f_i(\mathbf{x}i^L, \mathbf{x}{i+1}^L; \mathbf{o}_i^L)$ for GPS/INS and LiDAR respectively.
\par
By using the GPS/INS-based pose as our main reference, our concept involves incorporating racing line waypoints, given that race tracks typically have a loop-shaped configuration. 
Therefore, we propose a hash-based racetrack ID matching algorithm to associate a racetrack ID with GPS pose keys and corresponding LiDAR pose keys.
For each GPS/INS pose key $\mathbf{x}_i^G$ and corresponding LiDAR pose key $\mathbf{x}_i^L$:
\[
\text{racetrack\_dict}[\mathbf{x}_i^G, \mathbf{x}_i^L] \rightarrow \textit{racetrack\_geometric\_id} .
\]
\par
After successfully finding a loop closing pose from corresponding \textit{racetrack\_geometric\_id}, we update the GPS/INS-based and LiDAR-based pose factors as follows:
\begin{equation}
\begin{alignedat}{2}
& f_i(\mathcal{X}_i^G) = f_i(\mathbf{x}_i^G, \mathbf{x}_{\text{loop}}^G; \mathbf{o}_{\text{loop}}^G) , \\
& f_i(\mathcal{X}_i^L) = f_i(\mathbf{x}_i^L, \mathbf{x}_{\text{loop}}^L; \mathbf{o}_{\text{loop}}^L) .
\end{alignedat}
\label{eq_loop_factor}
\end{equation}
\par
After conducting several iterative drives on the race course, we perform non-linear graph optimization using incremental smoothing and mapping(iSAM)-based optimization \cite{kaess2012isam2}. 
Subsequently, we run an iterative closest point (ICP) algorithm between the graphs to find a transformation matrix that aligns them. The transformation matrix is then used to generate a unified map, integrating the information from both GPS/INS-based and LiDAR-based pose graphs.

\subsection{Efficient LiDAR-based state estimator}
\subsubsection{Efficient registration method}
\label{sec:efficient_registration}
\begin{figure}[t]
    \centering
    \includegraphics[width=0.95\columnwidth]{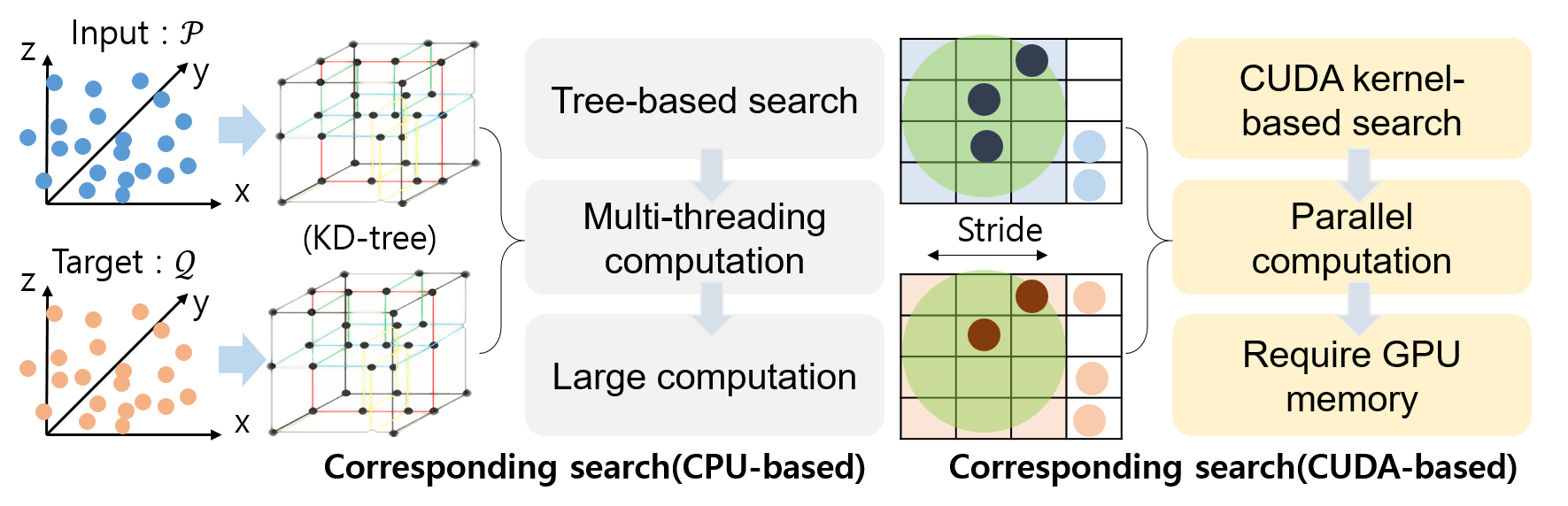}
    \caption[Comparison of Covariance Estimation Methods: CPU-Based vs. GPU-Based Approaches]{Comparison of Covariance Estimation Methods: CPU-Based vs. GPU-Based Approaches}
    \label{fig:Corresponding_cpu_cuda}
\end{figure}

We employ the generalized iterative closest point (GICP) variant algorithm \cite{segal2009generalized, koide2021voxelized}, to employ the scan match algorithm by modeling the consequtive scans $\mathbf{P} = \{\mathbf{p}_{0}, \mathbf{p}_{1}, ..., \mathbf{p}_{n}\}$,  $\mathbf{Q} = \{\mathbf{q}_{0}, \mathbf{q}_{1}, ..., \mathbf{q}_{m}\}$ as a Gaussian distribution $\mathbf{p}_{i} \sim \mathcal{N} (\hat{\mathbf{p}_i}, C^{\mathbf{p}}_i)$,  $\mathbf{q}_{i} \sim \mathcal{N} (\hat{\mathbf{q}_i}, C^{\mathbf{q}}_i)$. 
Subsequently, the transformation error $d_i$ can be defined as
\begin{equation}
\begin{aligned}
    {d_i} = \hat{\mathbf{q}_i} - \mathbf{T}\hat{\mathbf{p}_i}.
\end{aligned}
\label{eq_trans_err}
\end{equation}
Thus, the $d_i$ distribution can be expressed as
\begin{equation}
\begin{aligned}
    {d_i} &\sim \mathcal{N} (\hat{\mathbf{q}_i} - \mathbf{T}\hat{\mathbf{p}_i}, C^{\mathbf{q}}_i - \mathbf{T}^TC^{\mathbf{p}}_i\mathbf{T}) \\
    &= \mathcal{N} (0, C^{\mathbf{q}}_i - \mathbf{T}^TC^{\mathbf{p}}_i\mathbf{T}).
\end{aligned}
\label{eq_trans_err_dist}
\end{equation}
Therefore, the registration problem can be represented as minimizing the error between  $\mathbf{P}$ and  $\mathbf{Q}$ as
\begin{equation}
\begin{aligned}
    \mathbf{T} =\arg\min_{\mathbf{T}}(\sum_{i}d_i^T(C^{\mathbf{q}}_i - \mathbf{T}^TC^{\mathbf{p}}_i\mathbf{T})d_i).
\end{aligned}
\label{eq_trans_likelihood}
\end{equation}
Here we found that CPU-based registration may encounter bottlenecks due to its sequential nature. 
More precisely, finding corresponding points and computing covariance when estimate the $C^{\mathbf{p}}_i$ and $C^{\mathbf{q}}_i$ for relevant points can be computationally expensive, leading to slower performance, particularly with large and complex point clouds as depicted in Fig. \ref{fig:Corresponding_cpu_cuda}.
\par
By adopting GPU-based methods \cite{koide2021voxelized}, we can overcome these bottlenecks and achieve faster and more scalable point cloud registration. 
By implementing a GPU-based nearest points search and covariance computation using the kernel descriptors, we can leverage parallel processing capabilities to significantly accelerate the registration algorithm. 
GPU computation is ideal for handling large-scale point cloud data, enabling efficient computations on multiple points simultaneously.
\par
Here, we consider several kernel descriptors, each with its corresponding equation, to measure similarity between data points.
The Radial basis function (RBF) kernel ($K_{\text{rbf}}$) \cite{koide2021voxelized}, The Gaussian kernel ($K_{\text{gauss}}$), the Polynomial kernel ($K_{\text{poly}}$), the Histogram Intersection kernel (HI) ($K_{\text{hist}}$), and the Laplacian kernel ($K_{\text{laplacian}}$) are presented in Table \ref{tab:kernel_descriptors}. 
In this study, the Laplacian method demonstrates the best performance among the considered kernel descriptors in GPU-based registration methods.
Our implementation extends and optimizes the Voxelized-GICP algorithms, resulting in improved results. 
\begin{table}[!tbp]
    \centering
    \caption{Kernel Descriptor for GPU-based covariance estimation}
    \label{tab:kernel_descriptors}
    \begin{tabular}{l|l}
    \hline
    \textbf{Kernel Descriptor}       & \textbf{Equation}\\
    \hline    
    RBF              & $K_{\text{rbf}}(x, y) = \exp\left(-||x - y||^2 * \sigma\right)$ \\ \hline
    Gaussian              & $K_{\text{gauss}}(x, y) = \exp\left(-\frac{||x - y||^2}{2\sigma^2}\right)$ \\ \hline
    Polynomial       & $K_{\text{poly}}(x, y) = (\alpha \langle x, y \rangle + c)^d$           \\ \hline
    HI & $K_{\text{hist}}(x, y) = \frac{\sum \min(x[i], y[i])}{\sum x[i]}$    \\ \hline
    Laplacian        & $K_{\text{laplacian}}(x, y) = \exp\left(-\frac{||x - y||}{\sigma}\right)$ \\
    \hline
    \end{tabular} \\
\end{table}

\subsubsection{LiDAR-inertial-vehicle model(LIV) odometry for prediction}
We propose an efficient LIV odometry algorithm using LiDAR, IMU, and vehicle state data, including velocity $\hat{\mathbf{v}}_t$ and wheel steering angle $\hat{\delta}_t$ to address high-speed challenges.
This fusion of inertial information and vehicle state data ensures precise, real-time pose predictions, even in the middle of prevalent slips and vibrations in high-speed driving.
By utilizing LIV odometry predictions for the initial transformation, we aim to enhance the convergence and accuracy of the GICP-variant algorithm during scan matching providing a more informed starting point for the iterative scan matching process.
\par
Firstly, we incorporate inertial data, specifically the rotational velocity $\hat{\mathbf{w}}_t = \mathbf{w}_t + \mathbf{b}_t^w + \mathbf{n}_t^w$ where $\mathbf{b}_t^w$ and $\mathbf{n}_t^w$ are bias and noise, respectively \cite{forster2016manifold}.
For rotational pose prediction, we use the quaternion $\Delta\mathbf{q}_{ij}$ obtained from the IMU pre-integrator. 
In addition, we integrate the wheel steering angle $\hat{\delta}_t = \delta_t + b_t^{\delta} + n_t^{\delta}$ and velocity $\hat{\mathbf{v}}_t = \mathbf{v}_t + \mathbf{b}_t^{v} + \mathbf{n}_t^{v}$ to apply . 
We predict the rotation matrix for the next time step as:
\begin{equation}
\begin{aligned}
\Delta\mathbf{q}_{ij} 
    &= \mathbf{q}_{i}^-1 \otimes \mathbf{q}_{j} + \mathbf{v}_{j} \frac{tan(\delta_j)}{L} \\
    &= \prod_{k=i}^{j-1}\left[\frac{1}{2} (\hat{\mathbf{w}}_t - \hat{\mathbf{w}}_k^w) + \mathbf{v}_{k}\frac{tan(\hat{\delta}_k - n_j^{\delta})}{L}\right]\Delta t,    
\end{aligned}
\end{equation}
where rotational pre-integration is computed by quaternion multiplication $\otimes$.
In addition, $\Delta t$ represents the precise time difference between IMU and vehicle information, with a interval of 0.01 seconds.
This approach allows us to estimate the rotational component of the pose even in the absence of direct inertial measurements.
\par
To address the noise inherent in the inertial sensor's acceleration measurements, we directly integrate $\hat{\mathbf{v}}_t$ for linear pose pre-integration to obtain $\Delta\mathbf{p}_{ij}$ as
\begin{equation}
\begin{aligned}
\Delta\mathbf{p}_{ij} 
    &= \prod_{k=i}^{j-1}\left[\mathbf{v}_k + \frac{1}{2} \mathbf{R}\Delta\mathbf{q}_k(\hat{\mathbf{v}}_k - \mathbf{b}_k^v) \right]\Delta t.    
\end{aligned}
\end{equation}
\par
We consider LiDAR odometry's next scan match initial guess $\mathbf{x}^{\dagger L}_{j}$ as : 
\begin{equation}
\mathbf{x}^{\dagger L}_{j} = \mathbf{x}^{L}_{i} *     
    \begin{bmatrix}
    \Delta \mathbf{q}_{ij} & \Delta \mathbf{p}_{ij} \\
    0 & 1
    \end{bmatrix}.
\end{equation}

\subsubsection{Scan match for correction}
Utilizing a unified frame map, we can leverage the scan match pose $\mathbf{y}^k_{scan} \in \mathbf{y}^k$ as a multimodal measurement. 
Our key idea is an efficient target map query process of the point cloud map $\mathbf{M}^W$ using a GPU-hash data structure. 
This process is vital, especially for large race tracks. 
By leveraging the GPU-hash data structure, we associate each pose on the race line with its corresponding point cloud map.
In addition, we employ a sliding-window approach by utilizing points from $\mathbf{M}_i^{\psi} \subset \mathbf{M}^W$ \cite{lee2021assistive, lee2022design} instead of processing the entire unified map $\mathbf{M}^W$. 
Furthermore, the ICP-variant algorithm's performance improves through predictive transformations of scan query data with accurate initial guessing predictions, $\mathbf{x}^{\dagger L}_{j}$.
\par
\begin{figure}[!t]
    \centering
    \includegraphics[width=0.9\columnwidth]{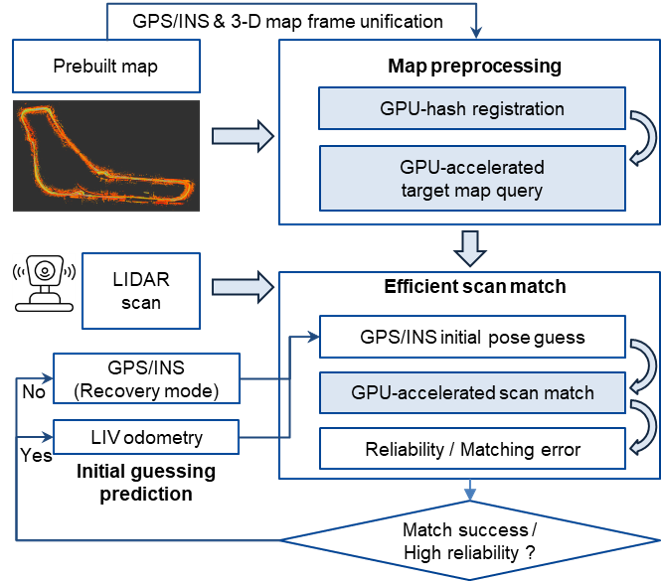}
    \caption[]{Proposed pipeline of GPU-accelerated scan match algorithm}
    \label{fig:scan_match}
\end{figure}

Finally, a reliability checker enables self-recovery for the LiDAR-based state estimator in case of scan matching failure or when reliability exceeds safety thresholds.
In such situations, we use the GPS/INS pose $\mathbf{x}_t^G$ for recovery, leveraging its georeferencing capabilities as the initial pose for scan matching.
A visual representation of our proposed pipeline is illustrated in Fig. \ref{fig:scan_match}.

\subsection{Wall detection for resilient navigation}
\label{sec:sub_wall_detection}
Since even the proposed robust localization system temporarily fails due to harsh conditions on the racetrack, we propose additional wall-following navigation module which does not depend on the main localization system. It offers enhanced resilience to the navigation system by providing safe lateral control in cases of complete localization system malfunction caused by multiple GPS failures.
To detect the walls of the racetrack at extreme driving speeds, we propose an efficient ground filtering algorithm that only uses direct LiDAR measurements.

When our proposed multimodal measurement fusion Kalman filter keeps computing $p(\mathbf{y}_{i}^{k}|\Theta^{k}, \mathbf{x}_{i})$ as $\lambda_{reject}$, we consider this status as positioning degraded situation.
To deal with these critical situations, we have designed our race car such that it follows along the race track wall, thereby avoiding collision with the wall.
To extract a wall in the racing track, we propose a wall detection algorithm comprising of fast ground removal and wall clustering algorithms. 
The algorithm helps navigate the wall resiliently when a deterioration is detected in the entire measurement.

\subsubsection{Vertical feature extraction for fast ground filtering}
\begin{figure*}[!t]
    \centering
    \subfigure[]{
        \includegraphics[width=0.63\columnwidth]{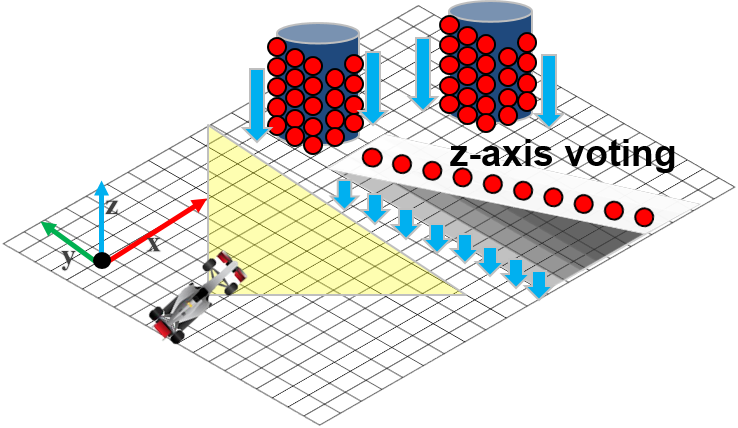}
        \label{fig:z_vote}
    }
    \subfigure[]{
        \includegraphics[width=0.63\columnwidth]{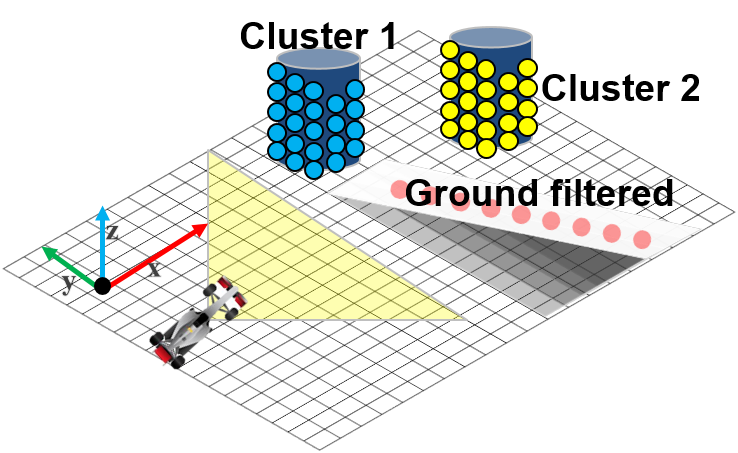}
        \label{fig:ground_filter}
    }
    \subfigure[]{
        \includegraphics[width=0.63\columnwidth]{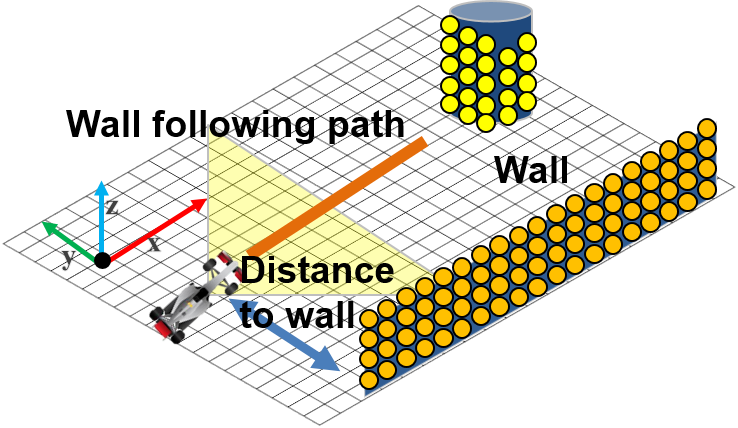}
        \label{fig:wall_detection}
    }    
    \label{fig:point}
    \caption[Description of process for ground points filtering, clustering, and wall detection]{
    Description of process for ground points filtering, clustering, and wall detection.
    (a) Ground filtering based on a z-axis voting algorithm. The number of voting to grid-cell is compared to a hyper-parameter to extract the vertical features. In the illustration, points corresponding to the banked road have a few counts.  
    (b) Ground-filtered points are illustrated. In addition, the Euclidean-distance clustering algorithm is implemented.   
    (c) The length of clusters is used, assuming that the longest right-side cluster is the wall. Points of the wall are computed to estimate the wall curvature and distance to the wall.
    }
\end{figure*}

Generally, the race track is mostly banked road, which angles towards the race track center to help vehicles speed up at curved corners.
Thus, the ground filtering algorithm for LiDAR points has to consider the road gradient. \\
We propose a novel vertical feature extractor using a hashing algorithm to consider the sparsity of LiDAR point-cloud data. 
Let $B \subset \mathbb{R}^3$ define the vehicle body coordinates.
Here, $B$ annotated values indicating the information obtained from the vehicle body’s origin---i.e., the center point of the rear axle. 
We also define the voxel-filtered LiDAR points $\mathbf{p}^B_{t}= \{p_1^B, \dots, p_k^B\}$ at time $t$, where $p_i^B$ is a voxelized point from the incoming LiDAR points.
After voxelization, we project $\mathbf{p}^B_{t}$ to the 2-D grid to vote the points corresponding to the grid-cell, as shown in Fig. \ref{fig:z_vote}.
Moreover, we propose a hashing algorithm during the voting to account for the sparsity of the point cloud.
\begin{equation}
\begin{aligned}
    & H(p_i^B) = index, \\
    & H^{-1}(index) = p_i^B,
\end{aligned}
\label{eq_hash_table}
\end{equation}
where hash function $H(x)$ maps the value $x$ at the table.
We iterate voxelized points $\mathbf{p}^B_{t}$ to vote on the corresponding grid-cell using $f(p_i^B) = g(u_i, v_i, \mathbf{n}_{index})$ where $\mathbf{n}_{index}$ contains the number of points voted and its hashing-index.
Thus, by comparing $\mathbf{n}_{index}$ size with hyper-parameter, we can efficiently filter out ground points $\mathbf{p}^{ground}_{t}$ from $\mathbf{p}^B_{t}$ in real-time without matrix computation for plane extraction.
\subsubsection{Wall following navigation}
To extract the plane feature from the point-cloud, we implement a random sample consensus (RANSAC)-variant algorithm \cite{qian2014ncc, li2017improved}.
However, a distance parameter determines the inlier or outlier points, which is unsuitable for curved wall areas.
Our approach uses a common tree-based Euclidean distance algorithm to find a wall from the ground-filtered points $\mathbf{p}^{filtered}_{t}$, as shown in Fig. \ref{fig:wall_detection}.
We implement a CUDA-based Euclidean distance clustering algorithm \cite{karbhari2018gpu, nguyen2020fast} to find cluster $C_j$ \cite{rusu2010semantic}:
\begin{equation}
\begin{aligned}
    & C_j = \arg \min_{i}\parallel{p}^{filtered}_i - \mu_j\parallel_{2}, \\
    & \mu_{j} = \frac{\sum_{i=1}^{m}1\{c^i=j\}{p}^{filtered}_i}{\sum_{i=1}^{m}1\{c^i=j\}},
\end{aligned}
\label{eq_clustering}
\end{equation}
where Eq. \ref{eq_clustering} is the subject condition for clustering algorithm.
We assume the longest cluster as wall cluster $\mathbf{w}_i(x_{w,i},y_{w,i})$ :
\begin{equation}
\begin{aligned}
    \mathbf{w}_i(x,y) = \mathbf{P}_z \cdot \arg\max_{x}(\mathcal{H}(C_{j}(x_{w,i},y_{w,i},z_{w,i}))),
\end{aligned}
\label{eq_wall_extract}
\end{equation}
where $\mathcal{H}(x)$ extracts the length of clusters, and $\mathbf{P}_z$ is the z-directional projection matrix. \\
We then obtain a coefficient of the polynomial regression model $\hat{\bm{\beta}}_w$ of $\mathbf{w}_i(x_{w,i},y_{w,i})$ as follows:
\begin{equation}
\begin{aligned}
    \hat{\bm{\beta}}_w = (\mathbf{X}_{w}^{T}\mathbf{X}_{w})^{-1}\mathbf{X}_{w}^{T}\mathbf{y}_{w},
\end{aligned}
\label{eq_wall_find}
\end{equation}
where $\mathbf{X}_{w} = \{ x_{w,i} \in \mathbf{w}_i(x_{w,i},y_{w,i})\}_{i = 1:n}$ and $\mathbf{y}_{w} = \{ y_{w,i} \in \mathbf{w}_i(x_{w,i},y_{w,i})\}_{i = 1:n}$.
We next estimate the distance to the wall used for collision warning as follows:
\begin{equation}
\begin{aligned}
    d_w = \mathbf{y}_{w}(x_{w,0}),
\end{aligned}
\label{eq_wall_distance}
\end{equation}
where $x_{w,0}$ denotes the origin point on body coordinates of the vehicle's $x$-directional axis. Thus, we can calculate the desired lateral shift from the current position, $d_{w} - d_{gap}$, where $d_{gap}$ denotes the desired distance from wall. Subsequently, we are able to generate a wall following path $\mathbf{p}_{w}$ that maintains a desired distance from the wall as
\begin{equation}
\begin{aligned}
    \mathbf{p}_w(x_{w,i}) = \hat{\beta}_2 {x_{w,i}^2} + \hat{\beta}_1 { x_{w,i}+ (d_{w} - d_{gap})},
\end{aligned}
\label{eq_wall_path}
\end{equation}
where $\{\hat{\beta}_1, \hat{\beta}_2 \} \subset \hat{\bm{\beta}}_w$.
The path is generated using the second-order polynomial curve fitting to address the characteristics of oval tracks. 
Afterwards, the generated path is being handled by the existing controller to track the given path accordingly.
More details about the controller can be found in our other study \cite{seong2023data}.


\section{Results}
\label{sec:results}
\subsection{Test scenarios}
During our evaluation, we conducted tests at speeds of up to 248.8 $km/h$ across four different race events held in different countries. The maximum speeds achieved for each event are detailed in Table \ref{tab.field_summary}.
\par
The US tracks, including Lucas Oil Raceway, Indianapolis Motor Speedway, Las Vegas Motor Speedway, and Texas Motor Speedway, provided a suitable environment to test high-speed scenarios. On the other hand, the Monza Circuit in Italy and Everland Speedway in South Korea offered more dynamic scenarios with challenging corners such as hairpin turns and chicanes. These dynamic features posed significant challenges for state estimation, especially in dealing with lateral derivatives and dynamic rotational driving.
\par

\begin{figure}[!hbt]
    \centering
    \subfigure[]{
        \includegraphics[width=1.0\columnwidth]{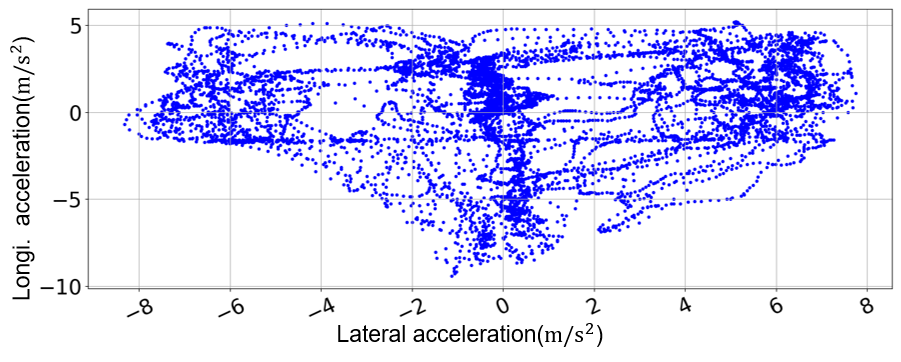}
        \label{fig:test_scenarios_hyundai}
    }
    \subfigure[]{
        \includegraphics[width=1.0\columnwidth]{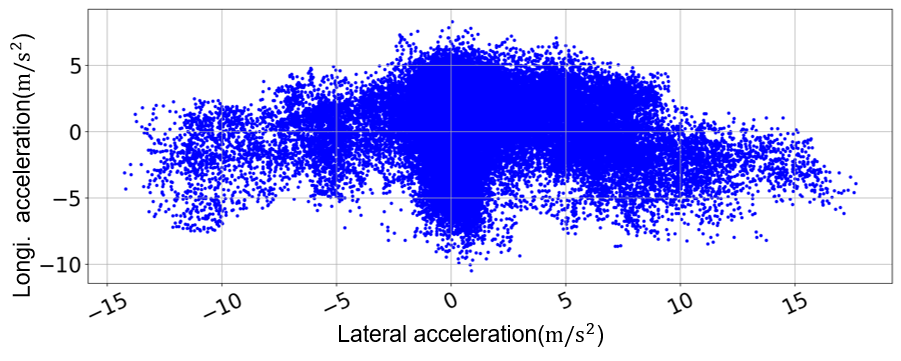}
        \label{fig:test_scenarios_monza}
    }    
    \caption[GG-diagram illustrating different test scenarios]{
    GG-diagram illustrating different test scenarios.
    Longitudinal and lateral acceleration data are collected using IMU sensors.
    (a) GG-diagram of Everland Speedway, driven by Hyundai IONIQ 5.
    (b) GG-diagram of Monza Circuit, driven by Dallara Indy Lights car.
    }
    \label{fig:test_scenarios}
\end{figure}

Figure \ref{fig:test_scenarios} depicts the gg-diagram of the Monza and Everland circuits, offering insights into the test scenarios by plotting longitudinal acceleration against lateral acceleration.
\begin{table}[h]
\centering
\caption[Field tests performance summary.]{Field tests performance summary.}
\label{tab.field_summary}
\begin{tabular}{lcccccc}
\hline
\multicolumn{1}{l|}{\textbf{Track}} & \textbf{LOR} & \textbf{IMS} & \textbf{LVMS} & \textbf{TMS} & \textbf{MC} & \textbf{ES}    \\ \hline
\multicolumn{1}{l|}{\begin{tabular}[c]{@{}l@{}}Maximum\\speed\end{tabular}} 
    & \begin{tabular}[c]{@{}l@{}}27.74 (m/s)\\ 99.9 (km/h)\end{tabular}
    & \begin{tabular}[c]{@{}l@{}}41.08 \\ 147.9 \end{tabular}
    & \begin{tabular}[c]{@{}l@{}}69.11 \\ 248.8 \end{tabular}
    & \begin{tabular}[c]{@{}l@{}}57.55 \\ 205.4 \end{tabular}
    & \begin{tabular}[c]{@{}l@{}}55.72 \\ 200.6 \end{tabular}
    & \begin{tabular}[c]{@{}l@{}}42.04 \\ 151.34 \end{tabular}
\end{tabular}
\\
{\raggedright 
    \textbf{LOR} : Lucas Oil Raceway, US
    \textbf{IMS} : Indianapolis Motor Speedway, US \\
    \textbf{LVMS} : Las Vegas Motor Speedway, US
    \textbf{TMS} : Texas Motor Speedway, US \\
    \textbf{MC} : Monza Circuit, Italy
    \textbf{ES} : Everland Speedway, South Korea
\par}
\end{table}

\subsection{Vehicle platform}
\subsubsection{Indy autonomous challenge}
The race car was based on a Dallara Indy Lights IL-15 chassis. The computation hardware was equipped with an ADLink x64 computer system based on an Intel Xeon with eight physical CPU cores and an NVIDIA RTX 8000 GPU.
The sensor unit has two independent NovAtel PwrPak7-Ds, with a full multi-frequency integrating GNSS engine and embedded micro-electro-mechanical system (MEMS) IMU, three Luminar LIDARs, three Aptiv radar, and six Allied Vision cameras. 
In addition, vehicle states, such as wheel speeds and brake pressures, are available.
More detailed information is represented in Table \ref{tab:platform}.
\begin{table}[!t]
\caption[Platform's computing and networking hardware specifications]{\textbf{Computing and networking hardware specifications}}
\label{tab:platform}
\begin{center}
    \begin{tabular}{l|l}
    Device   & Specification \\
    \hline\hline
    CPU            & Intel Xeon E 2278 GE – 3.30 GHz (16T, 8C)                                              \\\hline
    GPU            & Nvidia Quadro RTX 8000 x 1 (PCIe slot)                                                 \\\hline
    RAM            & 64 GB                                                                                  \\\hline
    Ethernet ports & 3 x GigE RJ45 (In-built)                                                               \\
                   & + 2 x 40GbE QSFP+ (PCIe-based)                                                         \\\hline
    CAN ports      & 4 x In-built ports, 2 x port on PCIe card                                              \\\hline
    Network switch & Cisco IE-5000-12s12p-10G -- 12 GigE copper                                             \\
                   & 12 GigE fiber, 4x 10G uplink, PTP GM Clock                                             \\\hline
    Wireless       & Cisco/Fluidmesh FM4500 - up to 500Mbp \\ \hline
    GPS            & 2 x NovAtel PwrPak7 + HxGN SmartNet RTK \\ \hline
    LiDAR          & 3 x Luminar Hydra 120$^{\circ}$  \\ \hline
    RGB camera     & 6 x optical camera \\ \hline 
    Radar          & 3 x Delphi(1 x ESR + 2 x MRR) 
    \end{tabular}
    \end{center}
\end{table}

\subsubsection{Hyundai Autonomous Challenge}
For the Hyundai competition, the race car was built on an Intel Xeon-based computation hardware with an NVIDIA GeForce RTX 3080 GPU. The vehicle was equipped with an array of sensors, including the Ouster OS2-128 LiDAR and two Velodyne VLP 16 puck LiDAR sensors mounted on the side parts. Additionally, the car featured four GMSL Seconix cameras for visual perception, while the GPS system consisted of a Ublox F9P receiver and a Microstrain 3DM-GQ7 INS for localization.
Detailed specifications of the computation and sensor hardware used in the Hyundai competition are provided in Table \ref{tab:hyundai_platform}.

\begin{table}[!t]
\caption[Hyundai competition's computing and sensor hardware specifications]{\textbf{Hyundai Competition's Computing and Sensor Hardware Specifications}}
\label{tab:hyundai_platform}
\begin{center}
    \begin{tabular}{l|l}
    Device & Specification \\
    \hline\hline
    CPU & Intel Xeon E 2278 G – 3.40 GHz (16T, 8C)           \\ \hline
    GPU & Nvidia GeForce RTX 3080 (PCIe slot) \\ \hline
    RAM & 64GB \\ \hline
    Ethernet ports & PCIe-PoE550X 2-port 10GbE \\ \hline
    CAN ports &  PEAK PCAN-PCI Express FD 2Ch\\ \hline
    Network switch & Tp-link TL-SX1008-10G \\ \hline
    Wireless & KT LTE Egg - up to 500Mbp \\ \hline
    GPS & Ublox F9P + Microstrain 3DM-GQ7 \\ \hline
    RTK & Network RTK + Synerex MRD-1000V2 \\ \hline
    LiDAR & Ouster OS2-128 + 2 x Velodyne VLP 16 \\ \hline
    RGB camera & 4 x GMSL Seconix camera \\ \hline
    \end{tabular}
    \end{center}
\end{table}

\subsection{Multimodal measurement fusion Kalman filter}

\subsubsection{Resilience to simulated error and noise}
\begin{figure}[!t]
    \centering
    \subfigure[]{
        \includegraphics[width=0.8\columnwidth]{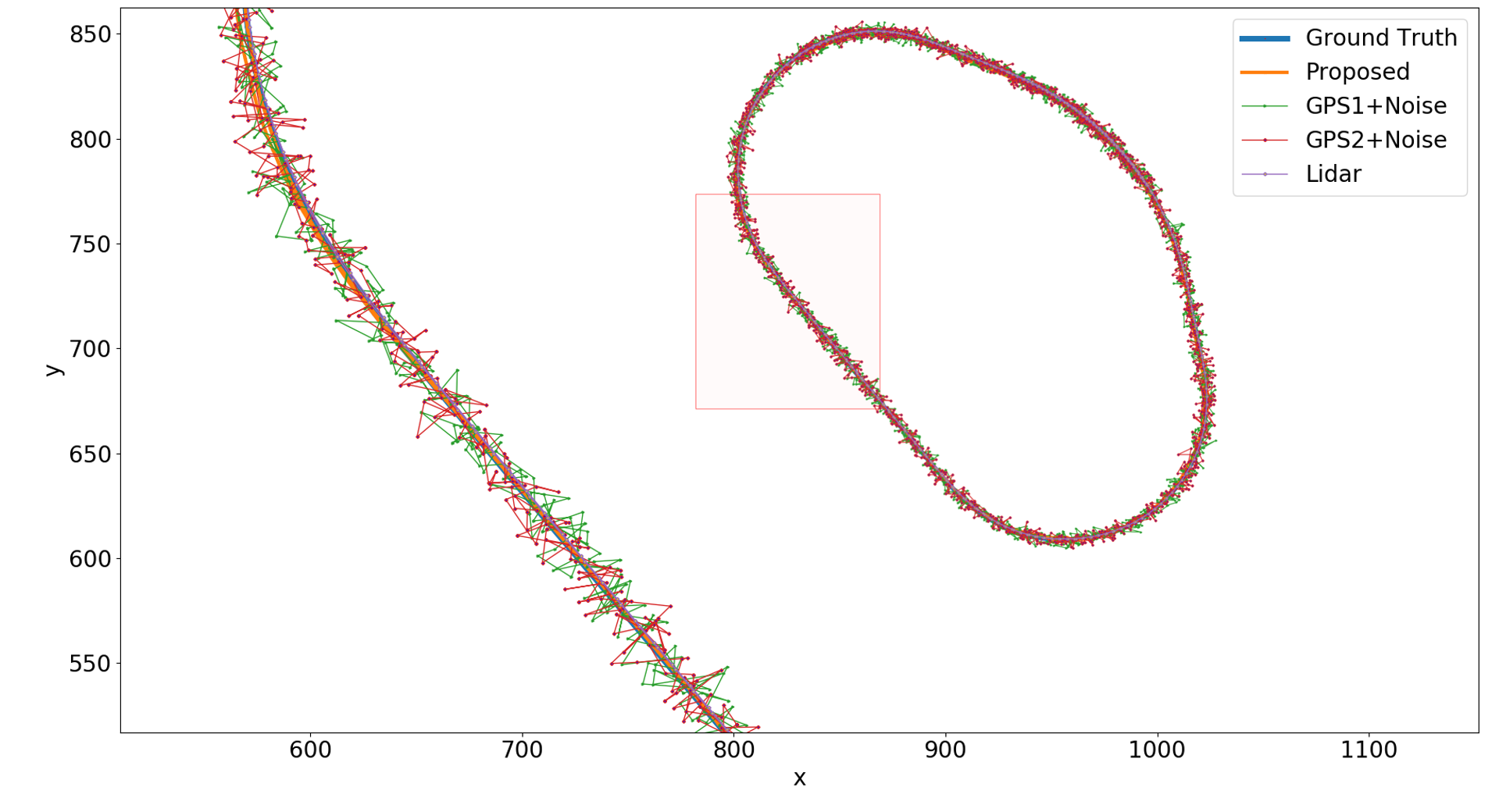}
        \label{fig:both}
    }
    \subfigure[]{
        \includegraphics[width=0.8\columnwidth]{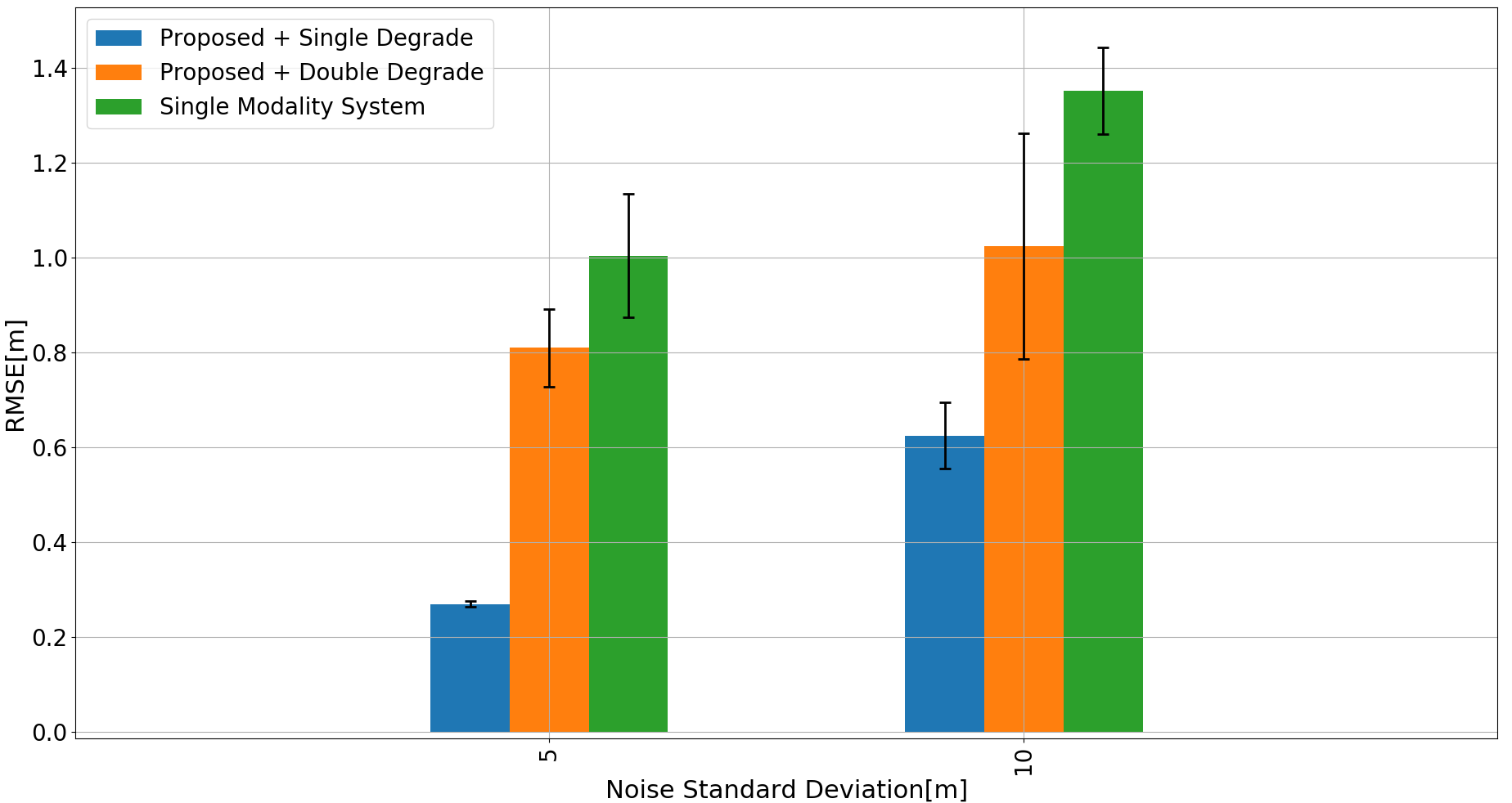}
        \label{fig:max_min}
    }    
    \label{fig:result_simulation}
    \caption[This should be changed to clear plot.]{
    Large-noise was artificially added to the multimodal measurements and the single modality system.
    It is crucial to note that the artificial noise does not accurately replicate the intricacies of real-world GPS noise; it serves as an approximation and should not be conflated with actual GPS noise.
    (a) 
    A simulated error on the LVMS race track. 
    The degradation of both GPS is represented in magnified view. 
    (b) Proposed algorithm successfully estimates the position robustly, compared to single modality system.
    Specifically, the accuracy performance was maintained even when one of the two degradation occurred.
    }
\end{figure}
We performed simulated tests to evaluate the capability of the proposed algorithm.
We introduced significant artificial noise into the recorded multimodal data, which included measurements from two GPS and one LiDAR sensors. This artificial noise was generated with a standard deviation of approximately 5m and 10m, as illustrated in Fig. \ref{fig:both}. It is important to clarify that the artificial noise, denoted as $\sigma_{sim}$, is employed to assess the system's robustness. However, it is crucial to note that $\sigma_{sim}$ does not accurately replicate the intricacies of real-world GPS noise; it serves as an approximation and should not be conflated with actual GPS noise.
\par
During the evaluation scenario, the vehicle averaged 225 $km/h$ and reached 244 $km/h$ at maximum speed. 
We compared the estimated pose error with the ground truth under conditions where one or two of the multimodal measurements were degraded, as well as in a single-modality system under identical conditions.
When compared to a system based on a single modality, the proposed algorithm successfully estimates the position robustly. 
Specifically, the accuracy performance was maintained even when one of the two degradation occurred as depicted in Fig. \ref{fig:max_min}.

\subsubsection{Real-world test}

\begin{figure}[!ht]
    \centering
    \subfigure[]{
        \includegraphics[width=0.45\columnwidth]{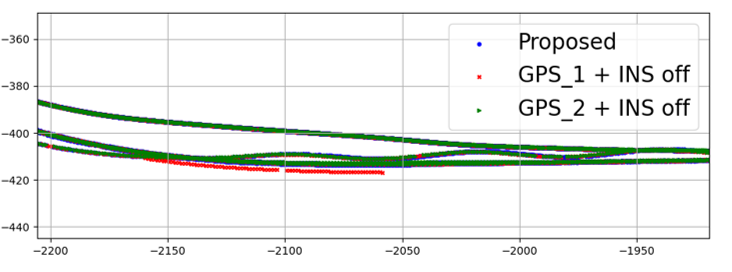}
        \label{fig:localize_sub_1}
    }
    \subfigure[]{
        \includegraphics[width=0.45\columnwidth]{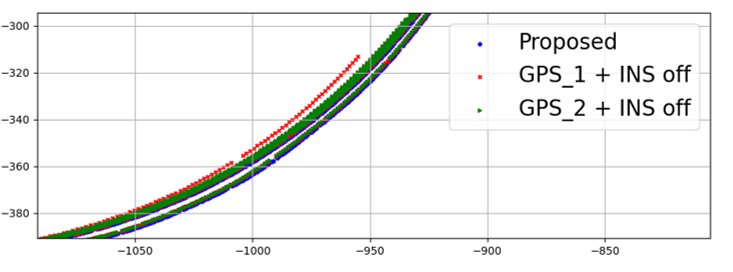}
        \label{fig:localize_sub_2}
    }
    \subfigure[]{
        \includegraphics[width=0.45\columnwidth]{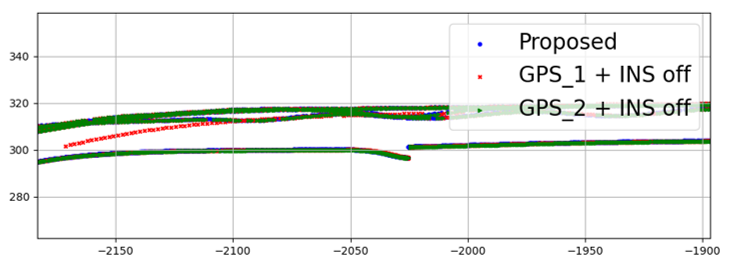}
        \label{fig:localize_sub_3}
    }    
    \subfigure[]{
        \includegraphics[width=0.45\columnwidth]{fig/result_localize_sub_3.png}
        \label{fig:localize_sub_4}
    }     
    \subfigure[]{
        \includegraphics[width=0.90\columnwidth]{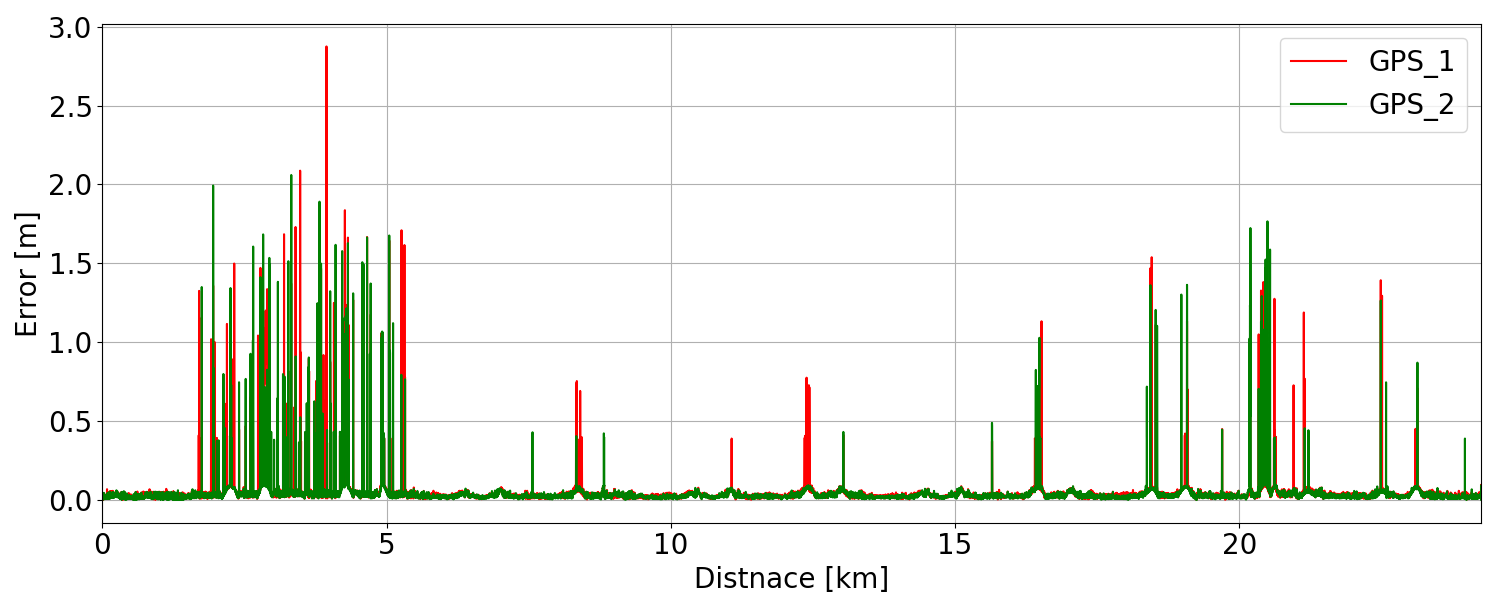}
        \label{fig:error_distance}
    }       
    \label{fig:result_localize}
    \caption[This should be changed to clear plot.]{
    Result of GPS log data and its estimated position using the proposed localization at IMS.
    Despite the degradation, the proposed localization algorithm estimated position resiliently throughout the track. 
    (a)-(d) A magnified representation of the track is shown with partial degradation of the GPS data and an estimated result.
    (e) Errors from two GPS units are shown during 20km driving. Initially, the GPS signal deteriorated frequently, and corner sections drifted frequently. Ground-truth is manually generated to compute error.
    }
\end{figure}

Regarding the specifics of implementation, we set $\epsilon$ as 0.2 $m$ and $\delta$ as 5.0 $m$ in Eq. \ref{eq_conditional_measure}.
Figures \ref{fig:localize_sub_1}-\ref{fig:localize_sub_4} present the magnified view of GPS and estimation data from the first day of the competition in IMS.
Specifically, on the first competition day, we experienced more difficulties with faultly GPS antennas and heavey vibration, leading to a degradation of GPS performance.
As depicted in Fig. \ref{fig:error_distance}, errors from two GPS units are shown during 20km driving.
Despite frequent degradation, we were able to effectively reject degraded measurements through the methods proposed.
\par
The operation of the tight INS/GNSS integration mode in the presence of heavy vibrations resulted in significant positioning degradation, particularly in the earlier IAC competitions. However, with the advancements in technology and the implementation of mechanical damping systems to mitigate vibrations, this issue has been gradually addressed and is now less prevalent in recent competitions.
\par
Figure \ref{fig:error_distance} shows the data from the first IAC competition in Indianapolis, illustrating the impact of vibrations on positioning accuracy.

\subsection{Unified frame map generation}
We evaluate the integrated robust state estimator, combining GPS/INS and efficient LiDAR-based state estimators, to assess its performance in high-speed and dynamic driving scenarios on the race track. 
Our evaluation aimed to validate the state estimator's accuracy and reliability under challenging conditions.
\par
Additionally, we introduced simulated LiDAR point cloud noise to create scenarios where the scan matching algorithm might fail during the drive. This allowed us to assess the robustness of the integrated system and evaluate its ability to recover the LiDAR-based state estimation using the GPS/INS modality as a reliable fallback.
A evaluation video can be found at the following link: \footnote{\href{https://youtu.be/qnaOq1lsJOo}{https://youtu.be/qnaOq1lsJOo}}.

\subsection{Efficient LiDAR-based state estimator}
We conducted a comprehensive evaluation using a dataset. 
Our LiDAR-based state estimator emphasizes that we can extend the capabilities of our high-speed racecar beyond reliance solely on GPS-based systems, even in GPS-denied environments such as overpasses and tunnels.
\subsubsection{Efficient registration method}
To enable dense registration using 128-channel or solid-state LiDAR, we encountered a significant CPU bottleneck during the covariance estimation process, particularly while performing the corresponding points search algorithm. 
To address this limitation and enhance computational efficiency, we successfully utilized CUDA programming, leveraging GPU acceleration for the nearest points search and covariance computation. 
This approach effectively overcame the performance constraints associated with CPU-based methods.
\par
Furthermore, we conducted an in-depth investigation into the impact of various kernel descriptors on the rotational error within our registration method. 
To ensure efficient computation, we systematically tested multiple kernel descriptors. 
Remarkably, our proposed Laplacian method emerged as the most effective, yielding the lowest errors across all evaluation metrics. Detailed results are presented in Table \ref{tab:kernel_descriptors_compare}.
\begin{table}[htbp]
    \centering
    \caption{Comprehensive Comparison of Kernel Descriptors}
    \label{tab:kernel_descriptors_compare}
    \begin{tabular}{l|cccc}
    \hline
    \textbf{Kernel Descriptor} & \textbf{rmse} & \textbf{mean} & \textbf{median} & \textbf{std}\\
    \hline    
    Polynomial & 164.93 & 143.97 & 125.01 & 80.47  \\ 
    HI  & 15.94 & 14.75 & 17.19 & 6.04  \\ 
    Gaussian   & 23.28 & 19.70 & 18.36 & 12.39\\ 
    RBF & 11.14 & 9.69 & 8.14 & 5.48 \\ 
    \textbf{Laplacian} & \textbf{10.67} &\textbf{ 9.24} & \textbf{8.17} & \textbf{5.32}\\
    \end{tabular} \\
    \begin{flushleft}
    \hfill * Leaf size = 1.0 m
    \end{flushleft}
\end{table}

\subsubsection{Evaluation on challenging dataset}
We conducted an evaluation of our efficient LIV odometry on the demanding Monza circuit to assess its performance in handling high lateral acceleration and dynamic rotation at speeds of up to 200 $km/h$ on the road course circuit.
\par
\begin{figure}[!hbt]
    \centering
    \subfigure[]{
        \includegraphics[width=0.9\columnwidth]{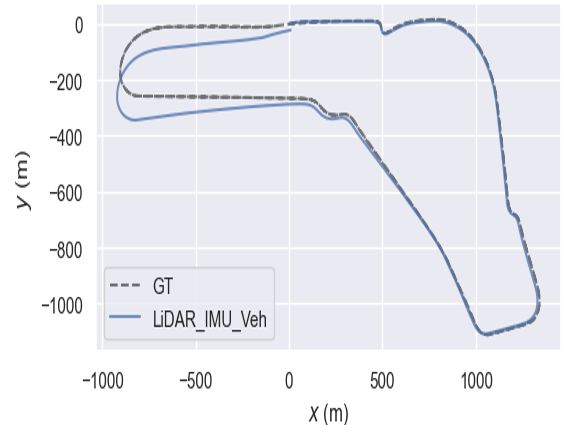}
        \label{fig:monza_lidar_veh_imu}
    }
    \subfigure[]{
        \includegraphics[width=0.9\columnwidth]{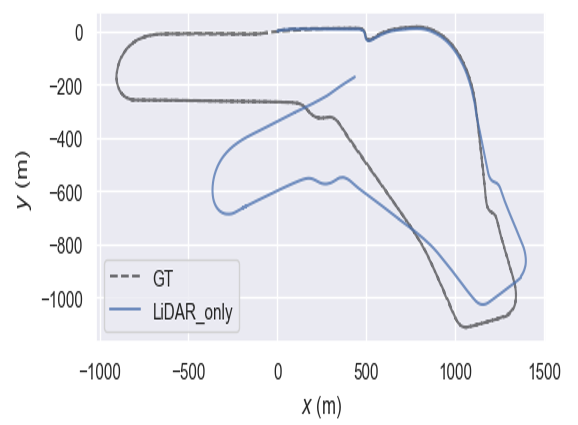}
        \label{fig:monza_lidar_only}
    }    
    \caption[Comparison of LiDAR Odometry with and without IMU and Vehicle Information]{
    Comparison of LiDAR odometry with and without IMU and vehicle information integration. 
    (a) LiDAR odometry with preintegrated IMU and vehicle information.
    (b) LiDAR-only odometry without preintegrated IMU and vehicle information.
    }
    \label{fig:monza_eval}
\end{figure}

When the inertial and vehicle models are not integrated in dynamic scenarios, our LiDAR-only odometry exhibits significant errors and struggles to cope with environments where similar features are repeated as shown in Fig \ref{fig:monza_eval}.
\par
To address this limitation, we integrated the LiDAR data with inertial and vehicle motion models, resulting in improved odometry performance and robustness in high-speed and dynamic driving scenarios. Our integrated approach effectively overcomes challenges posed by repeated features and yields more accurate and reliable state estimation, as demonstrated in the Monza circuit evaluation.
\par
A evaluation video can be found at the following link: \footnote{\href{https://youtu.be/C-H2i3tJvJ4}{https://youtu.be/C-H2i3tJvJ4}}.

\subsubsection{Comprehensive evaluation with benchmarks}
We conducted a comprehensive evaluation of our efficient LiDAR-based method by comparing it with publicly available LiDAR odometry algorithms. Our proposed cuda-accelerated method, which utilizes a pre-built map, demonstrated superior performance even with significantly lower CPU usage compared to other methods.
\par
While our proposed method may exhibit slightly less accuracy than methods using CPU-based covariance estimation, this discrepancy can be compensated for through pose graph optimization. Given that our deployment environment is a race track, we found that our loop closing method, which utilizes race track geometry, efficiently identifies re-visited positions and detects loops.
\par
\begin{figure*}[!hbt]
    \centering
    \subfigure[]{
        \includegraphics[width=0.62\columnwidth]{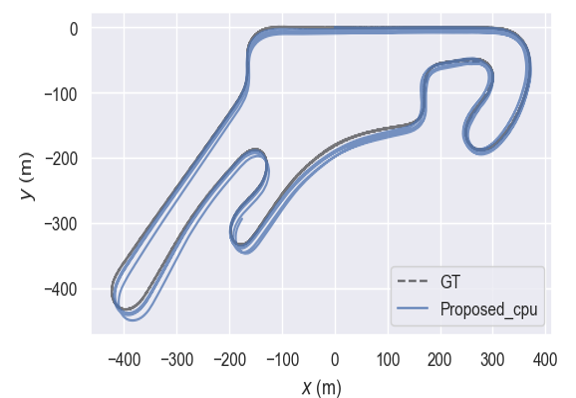}
        \label{fig:proposed_cpu}
    }
    \subfigure[]{
        \includegraphics[width=0.62\columnwidth]{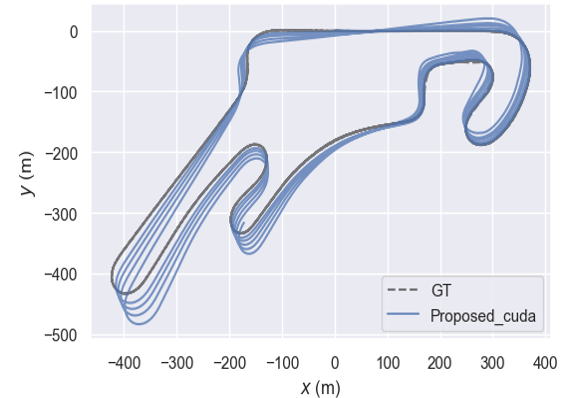}
        \label{fig:proposed_cuda}
    }
    \subfigure[]{
        \includegraphics[width=0.62\columnwidth]{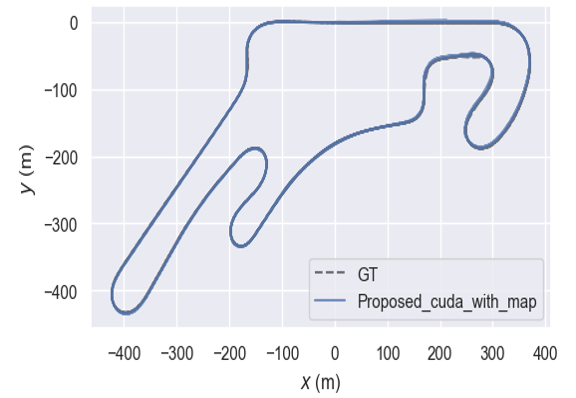}
        \label{fig:proposed_cuda_with_map}
    }
    \subfigure[]{
        \includegraphics[width=0.62\columnwidth]{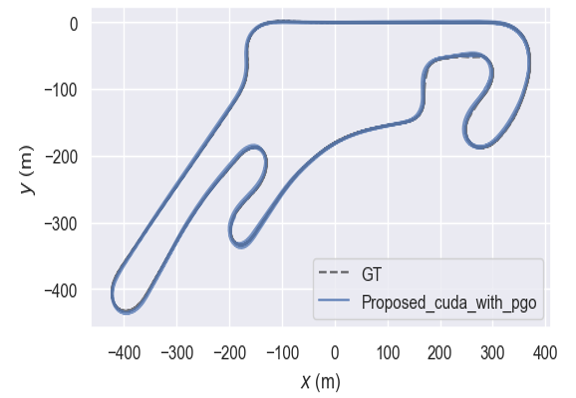}
        \label{fig:proposed_cuda_with_pgo}
    }
    \subfigure[]{
        \includegraphics[width=0.62\columnwidth]{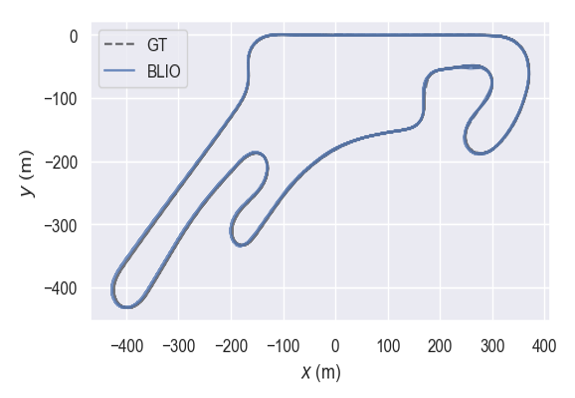}
        \label{fig:blio}
    }
    \subfigure[]{
        \includegraphics[width=0.62\columnwidth]{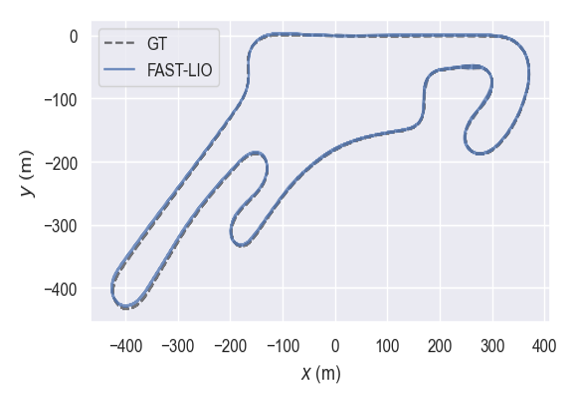}
        \label{fig:fast_lio}
    }
    \subfigure[]{
        \includegraphics[width=0.62\columnwidth]{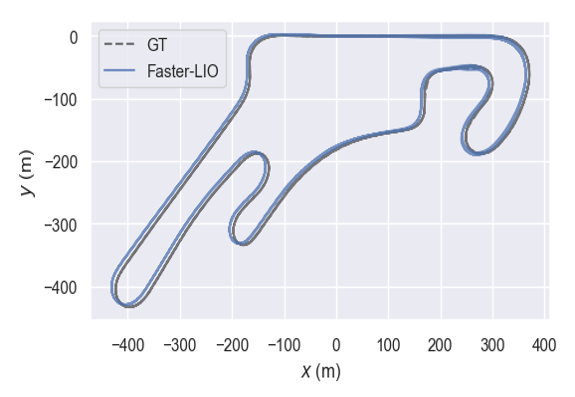}
        \label{fig:faster_lio}
    }
    \subfigure[]{
        \includegraphics[width=0.62\columnwidth]{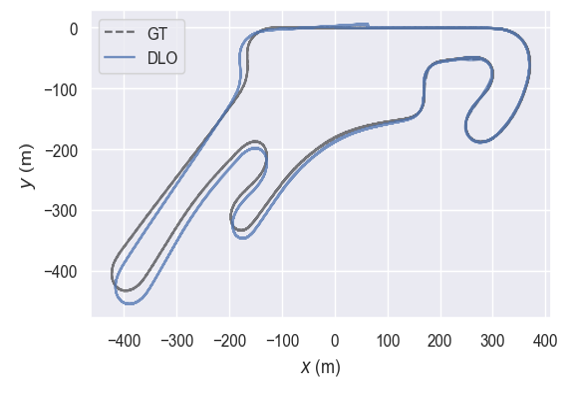}
        \label{fig:dlo}
    }
    \subfigure[]{
        \includegraphics[width=0.62\columnwidth]{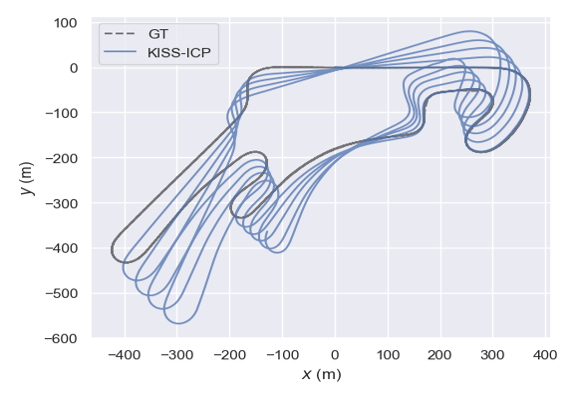}
        \label{fig:kiss_icp}
    }
    \subfigure[]{
        \includegraphics[width=1.95\columnwidth]{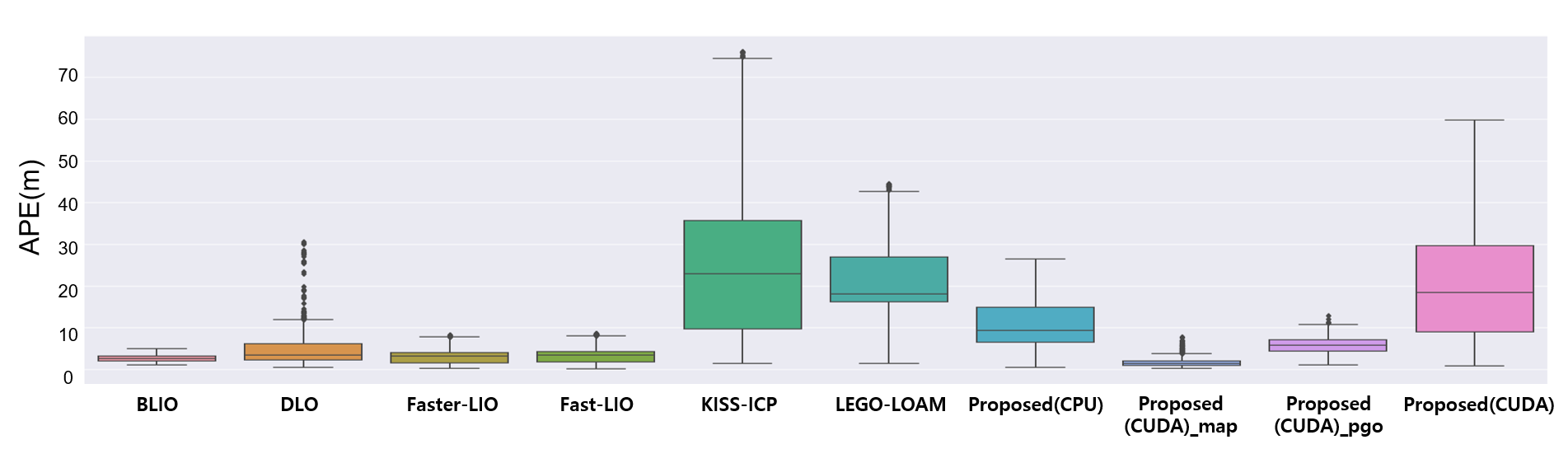}
        \label{fig:benchmark_hyundai_track}
    }
    \caption[Dallara AV-21 of Team KAIST at the Las Vegas Motor Speedway (LVMS).]{Comparison of different localization methods (a) V-FAST-GICP, (b) KISS-ICP, (c) GICP, (d) Proposed method. Ground truth is obtained by leveraging GPS signals during the map construction process
    (e) BLIO, (f) FAST-LIO, (g) Faster-LIO, (h) DLO, (i) KISS-ICP 
    }
    \label{fig:traj_result}
\end{figure*}

To further illustrate the effectiveness of our approach, we provide benchmark comparisons and present the results of LiDAR odometry evaluation on the Everland Speedway in Figure \ref{fig:traj_result} and Table \ref{tab:benchmark_result}, respectively. 
The data showcases the robustness and efficiency of our method in real-world scenarios.
\par
Furthermore, we emphasize the importance of managing computing resources. In situations where autonomous driving algorithms with substantial computing requirements are already running on a single PC, efficient deployment of the LiDAR-based approach without overloading the CPU is crucial.

\begin{table*}[htb!]
\caption[]{Comparison with the state-of-the-arts on ES dataset.}
\label{tab:benchmark_result}
\centering
\resizebox{0.9\textwidth}{!}{%
\begin{tabular}{l|cccc|ccc}
\multirow{2}{*}{Method} & \multirow{2}{*}{rmse} & \multirow{2}{*}{mean} & \multirow{2}{*}{median} & \multirow{2}{*}{std}  & \multicolumn{3}{c}{cpu(\%)} \\ \cline{6-8}
& & & & & \multicolumn{1}{c}{mean} & \multicolumn{1}{|c}{max} & \multicolumn{1}{|c}{min} \\
\hline  
BLIO \cite{kim2023adaptive} & 2.745 & 2.632 & 2.585 & \textbf{0.779} & 10.8 & 23.81 & 7.73 \\
DLO \cite{chen2022direct} & 6.142 & 4.775 & 3.414 & 3.862 & 3.22 & 4.92 & 2.24 \\
Faster-LIO \cite{bai2022faster} & 3.574 & 3.138 & 3.199 & 1.710 & 2.94 & 3.92 & 2.29 \\
Fast-LIO \cite{xu2021fast} & 3.801 & 3.319 & 3.430 & 1.853 & 2.45 & 2.97 & 2.13 \\
KISS-ICP \cite{vizzo2023kiss}& 29.802 & 24.655 & 22.907 & 16.74 & 8.28 & 12.79 & 5.63 \\
LEGO-LOAM \cite{shan2018lego} & 22.270 & 20.359 & 18.099 & 9.026 & 3.88 & 7.45 & 1.20 \\
Proposed(cpu) & 12.45 & 10.89 & 9.28 & 6.04 & 6.26 & 11.06 & 4.64 \\
Proposed(cuda) & 25.618 & 21.064 & 18.44 & 14.581 & \textbf{1.17} & \textbf{1.43} & \textbf{1.04} \\
Proposed(cuda)\_with\_map & \textbf{1.887} & \textbf{1.579} & \textbf{1.372} & 1.032 & 1.65 & 1.97 & 1.43 \\
Proposed(cuda)\_with\_pgo & 6.764 & 5.647 & 4.859 & 3.722 & 3.68 & 6.53 & 1.19 \\

\end{tabular}
 }
\begin{flushleft}
*Due to the high computational demands of DLO, KISS-ICP cannot be executed using a leaf size of 1.0m on dense point cloud data. Therefore, all algorithms were evaluated using data downsampled to 2.0m with identical conditions.
\end{flushleft}
\end{table*}

\subsection{Resilient navigation during GPS degradation}
\begin{figure}[!t]
    \centering
    \subfigure[]{
        \includegraphics[width=0.75\columnwidth]{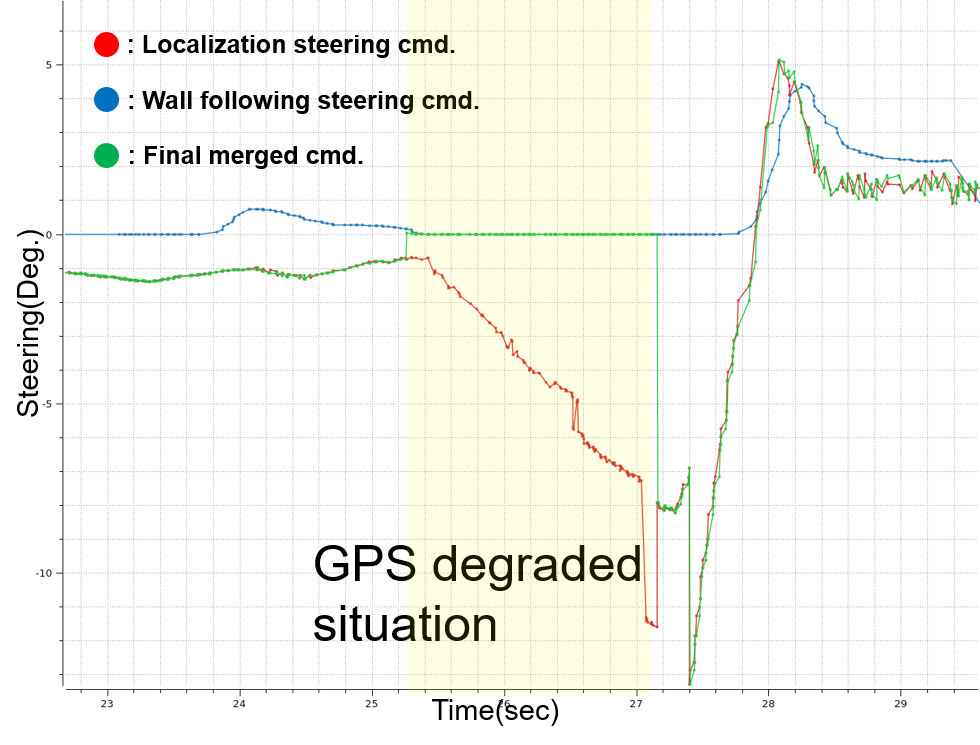}
        \label{fig:result_steering_1}
    }
    \subfigure[]{
        \includegraphics[width=0.75\columnwidth]{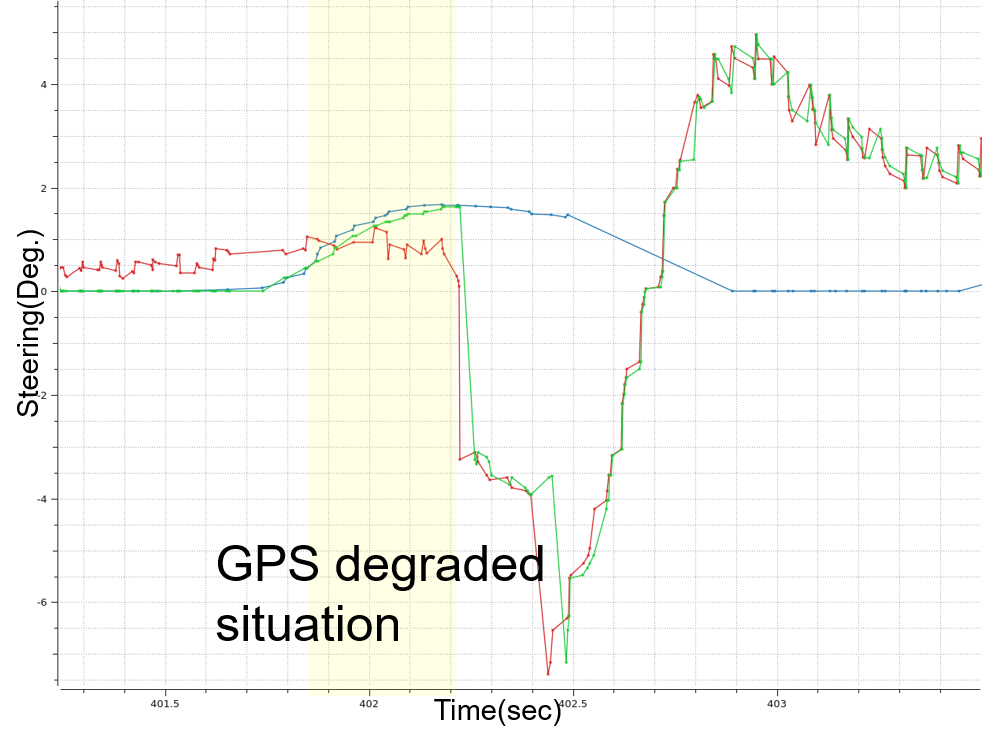}
        \label{fig:result_steering_2}
    }
    \label{fig:result_resilient_steer}
    \caption[This should be changed to clear plot.]{
    Proposed resilient navigation system which prevents crashes caused by critical GPS degradation. 
    (a) The proposed algorithm keeps the racing car at a certain distance from the wall for approximately 2 s to prevent crashing.
    (b) After receiving reasonable measurement data, the final steering command uses the localization-based command.
    }
\end{figure}

This study proposed a resilient navigation system that enables the race car to continue to follow the race track even in the event of a localization failure.
We computed wall following path $\mathbf{p}_w$ using Eq. \ref{eq_wall_path} utilizing direct perception information until the completion of localization recover. 
As shown in Fig. \ref{fig:result_steering_1},\ref{fig:result_steering_2}, our resilient navigation system came up with critical degradation on all GPS units.
In situations of localization failure caused by completely degraded measurement, our resilient navigation algorithm allowed the race car to maintain stability and continue driving away from wall.
When analyzing the lateral steering command that follows the racing line based on localization, we found the negative directional steering command causing the vehicle to crash onto the wall, as illustrated in \ref{fig:result_steering_1}.
More details can be seen online video \footnote{
\href{https://youtu.be/fiSqdMDmjGo}{https://youtu.be/fiSqdMDmjGo}}.

\subsection{Resilient navigation under incorrect desired path}
\begin{figure}[!t]
    \centering
    \subfigure[]{
        \includegraphics[width=0.65\columnwidth]{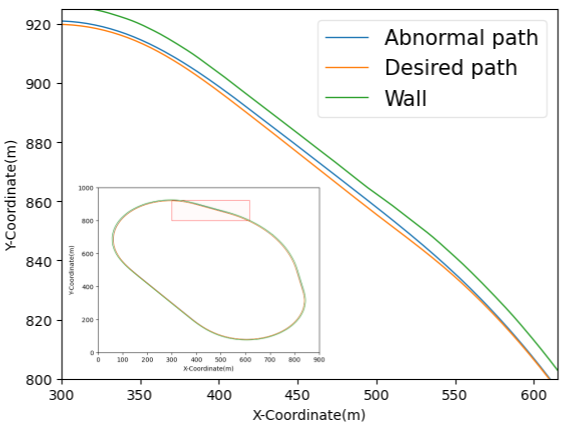}
        \label{fig:result_abnormal}
    }
    \subfigure[]{
        \includegraphics[width=0.65\columnwidth]{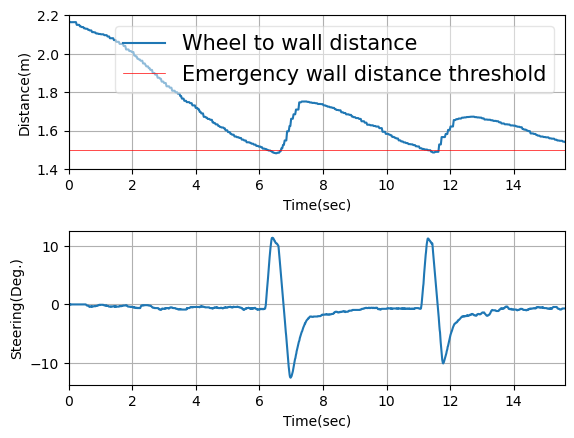}
        \label{fig:result_trigger}
    }
    \label{fig:result_resilient_abnormal_trigger}
    \caption[This should be changed to clear plot.]{
    Proposed resilient navigation system prevents abnormal cases from collisions. 
    (a)  A magnified representation of the red-shaded region of entire path. 
         An error parameter was present for the track boundary of the abnormal path,
         causing the wheel to wall distance abnormally small, that can potentially lead to collision with the wall.
    (b) The subplot below refers to the final merged command. The resilience system is triggered when the distance between the wheel and the wall falls below the established emergency threshold. 
        Positive steering values indicate counter-directional control against the wall.
    }
\end{figure}

During a test run, an incorrect desired path was accidentally given to the system. This happened due to the bias originated from the satellite imagery of the racetrack when the desired path was computed.  
As shown in Fig. \ref{fig:result_abnormal},  the incorrect desired path, also referred to as an abnormal path, could lead dangerously narrow gap to the race wall. 
When the distance between the wheel and wall became smaller than the emergency threshold as depicted in Fig. \ref{fig:result_trigger}, the resilient navigation system was triggered and provided counter-directional control. Thanks to the resilient navigation system, the wheel to wall distance was successfully maintained bigger than the emergency threshold 
 even though the vehicle kept trying to track the incorrect desired path and approaching to the race wall.

\section{Conclusion}
\label{sec:conclusion}
Our study introduces a comprehensive approach to enhance the state estimator for high-speed autonomous race cars, addressing challenges such as unreliable measurements, localization failures, and computing resource management. 
The proposed robust localization system utilizes Bayesian-based probabilistic evaluation of multimodal measurements, ensuring precise and reliable localization under harsh racing conditions. 
Additionally, a georeferenced LiDAR-based state estimator is seamlessly employed as a multimodal pose measurement, leveraging CUDA programming and GPU acceleration to overcome CPU bottlenecks. 
To handle potential localization failures, our resilient navigation system enables continuous track-following using direct perception information. 
Validation through real-world and simulation tests demonstrates robust performance, preventing accidents and ensuring race car safety.

\section{Discussion \& Future works}
\label{sec:discussion}
In this section, we delve into the limitations and concerns to offer detailed insights into real-world applications. Our study employs a multimodal measurement fusion method, emphasizing the prioritization of GPS/INS measurements. 
\par
Furthermore, in the generation of a unified frame map, we integrate RTK-GPS for centimeter-level accuracy. 
While our robust unified map generation framework, utilizing pose graph optimization, is generally resilient against typical noise, there is a concern regarding potential degradation in map quality when faced with significant noise.
\par
We designed a real-world test to operate the vehicle solely using the LiDAR-based state estimator without GPS/INS. However, further exploration is needed to assess the reliability of the LiDAR-only method, especially in high-speed scenarios. Therefore, in this study, we conducted a comprehensive evaluation using a dataset.
\par
As part of future work, we plan to explore advanced vision-based methods for drivable area perception, taking advantage of the evolving capabilities of edge computing. 
This exploration is motivated by a limitation in our current wall-following method, particularly when deploying on road-course tracks like Monza and Everland, where continuous walls are absent. 
In attempting to overcome this limitation, we experimented with a vision-based approach; however, a reliable and robust solution is yet to be established.

\section*{Acknowledgment}
We thank Dr Hoam Chung at Monash University for constructive feedback about the organization of the manuscript.


\bibliographystyle{unsrt}
\bibliography{citation.bib}

\begin{IEEEbiography}[{\includegraphics[width=1in,height=1.25in,clip,keepaspectratio]{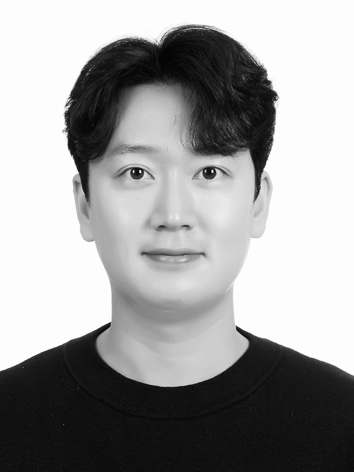}}]{Daegyu Lee} received his B.S. degree in Automotive Engineering from Kookmin University, Seoul, South Korea, in 2018, and earned his M.S. and Ph.D. degrees in the Division of Future Vehicle and Electrical Engineering from the Korea Advanced Institute of Science and Technology (KAIST), Daejeon, South Korea, in 2020 and 2024, respectively. 
He is currently a researcher at the Electronics and Telecommunications Research Institute (ETRI), Daejeon, South Korea.
Daegyu's research interests lie in autonomous systems and localization, focusing on robust and learning-based approaches.
\end{IEEEbiography}

\begin{IEEEbiography}[{\includegraphics[width=1in,height=1.25in,clip,keepaspectratio]{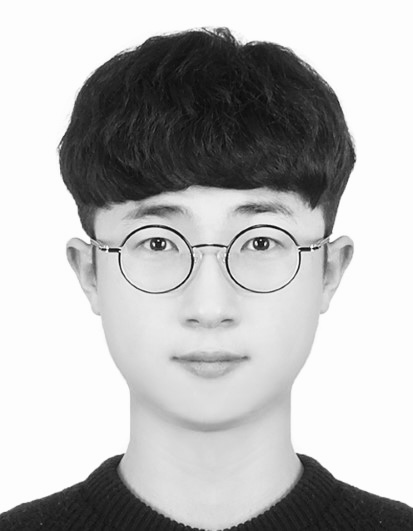}}]{Hyunwoo Nam} received his B.S. degree in Automobile and IT convergence, Kookmin University, Seoul, South Korea, in 2022, 
and his M.S. degree in Robotics Program from the Korea Advanced Institute of 
Science and Technology (KAIST), Daejeon, South Korea, in 2024. 
He is currently pursuing a Ph.D. degree in Electrical Engineering at KAIST.
His research interests include autonomous systems and localization
based on unmanned ground vehicles
\end{IEEEbiography}

\begin{IEEEbiography}[{\includegraphics[width=1in,height=1.25in,clip,keepaspectratio]{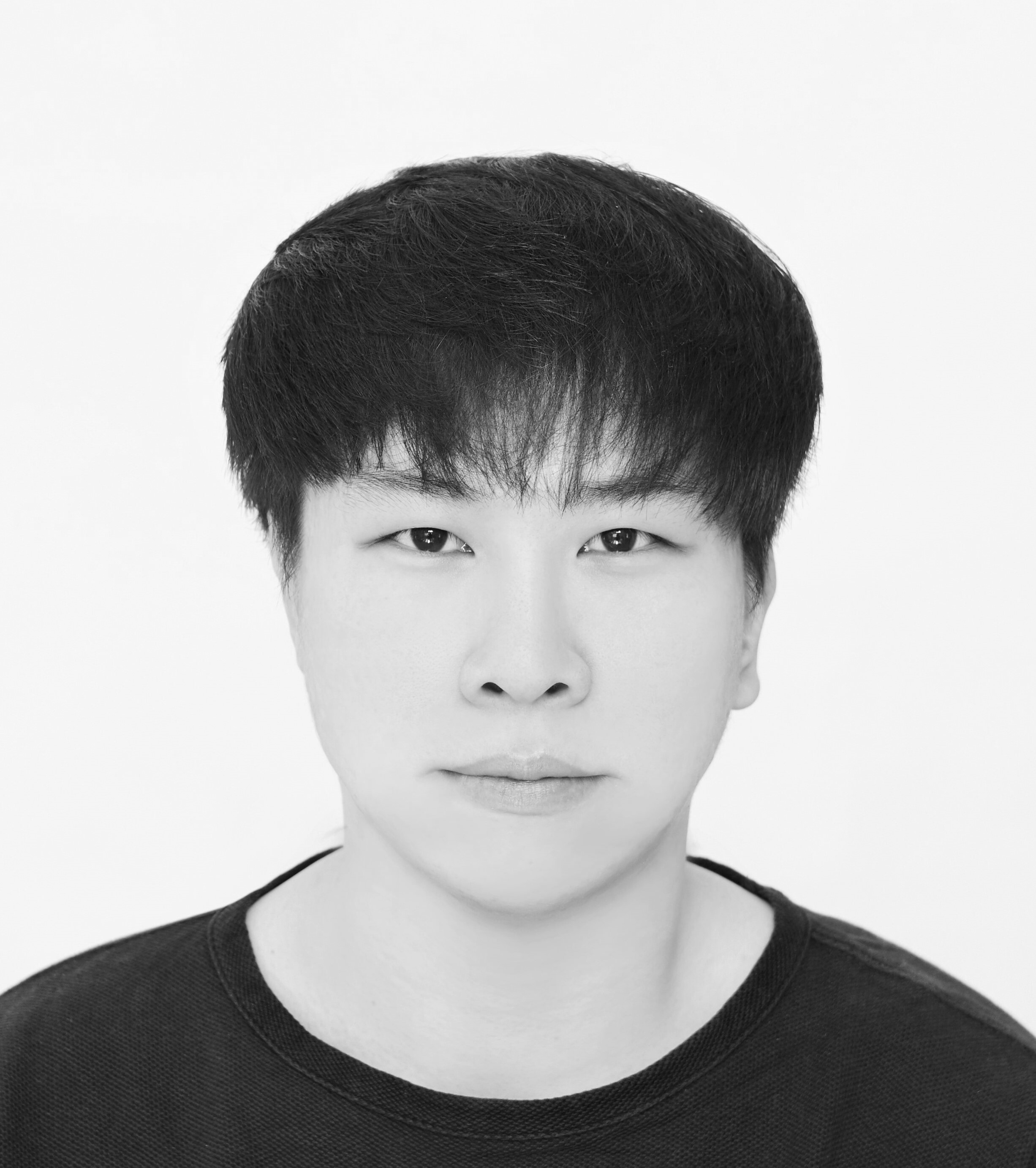}}]{Chanhoe Ryu} received his B.S. degree in Electrical Engineering, University of California at Los Angeles (UCLA), Los Angeles, CA, USA, in 2019, and he is currently pursuing a M.S. degree in Electrical Engineering from the Korea Advanced Institute of Science and Technology (KAIST), Daejeon, South Korea.
His research interests include autonomous systems, perception and motion planning on unmanned ground vehicles.
\end{IEEEbiography}

\begin{IEEEbiography}[{\includegraphics[width=1in,height=1.25in,clip,keepaspectratio]{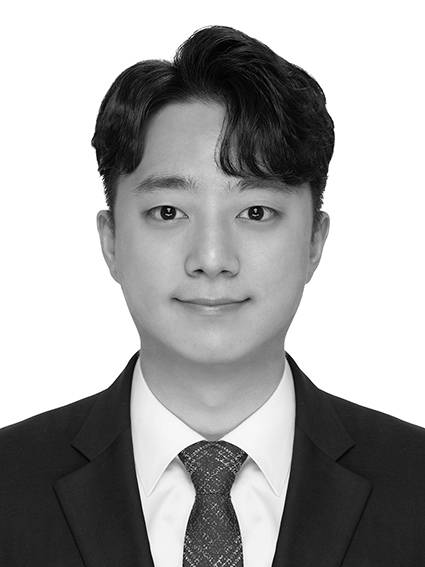}}]{Sungwon Nah} received his B.S. degree department of Mechanical Engineering from Konkuk University, Seoul, SouthKorea, in 2022, 
and his M.S. degree in Robotics Program from the Korea Advanced Institute of 
Science and Technology (KAIST), Daejeon, South Korea, in 2024. 
He is currently pursuing a Ph.D. degree in Electrical Engineering at KAIST.
His research interests include autonomous racing, 3D object detection, and trajectory prediction. 
\end{IEEEbiography}

\begin{IEEEbiography}[{\includegraphics[width=1in,height=1.25in,clip,keepaspectratio]{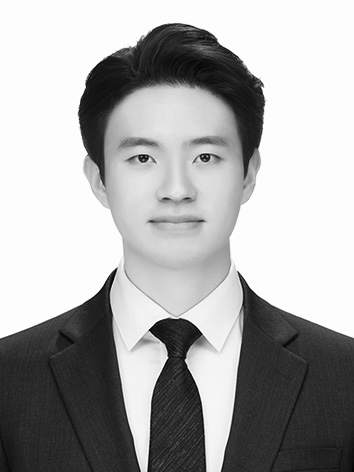}}]{Seongwoo Moon} received his B.S. degree in Electronic Engineering from Kyungpook National University, Daegu, South Korea, in 2022, and his M.S. degree in Electrical Engineering from the Korea Advanced Institute of 
Science and Technology (KAIST), Daejeon, South Korea, in 2024. 
He is currently pursuing a Ph.D. degree in Electrical Engineering at KAIST.
His research interests include robotics, motion planning, and control based on unmanned ground vehicles.
\end{IEEEbiography}

\begin{IEEEbiography}[{\includegraphics[width=1in,height=1.25in,clip,keepaspectratio]{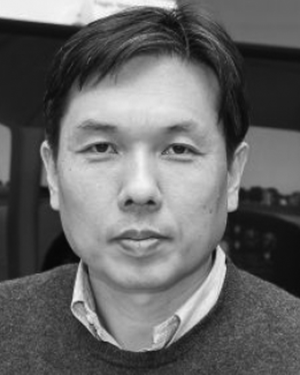}}]{D.Hyunchul Shim} received his B.S. and M.S. degrees in Mechanical Design and Production Engineering from Seoul National University, Seoul, South Korea, in 1991 and 1993, respectively, and his Ph.D. degree in Mechanical Engineering from the University of California at Berkeley, Berkeley, CA, USA, in 2000. He worked with the Hyundai Motor Company and Maxtor Corporation from 1993 to 1994 and from 2001 to 2005, respectively.
In 2007, he joined the Department of Aerospace Engineering, KAIST, Daejeon, South Korea, and is currently a tenured Professor with the Department of Electrical Engineering, and Adjunct Professor, Graduate School of AI, KAIST.
His research interests include control systems, autonomous vehicles, and robotics. He is also the Director of the Korea Civil RPAS Research Center.
\end{IEEEbiography}

\end{document}